\documentclass[journal,10pt]{IEEEtran}

\usepackage{ifpdf}
\usepackage{amsmath}
\usepackage{array}
\usepackage{fixltx2e}
\usepackage{stfloats}
\usepackage{url}

\usepackage{graphicx,psfrag,epsf}
\usepackage{mathtools}
\usepackage{booktabs}
\usepackage[english]{babel} 
\usepackage{amsmath,amsfonts,amsthm}
\usepackage{chngcntr}
\usepackage{ amssymb }
\usepackage{array}
\usepackage{booktabs} 
\usepackage{setspace}
\usepackage{fix-cm}
\usepackage{xcolor}
\usepackage[normalem]{ulem}
\usepackage{hyperref}
\hypersetup{colorlinks,urlcolor=blue}
\usepackage{cite}

\usepackage{url}
\usepackage{tikz}
\usepackage{cool}
\usepackage{enumitem}
\usepackage{soul}
\usepackage[normalem]{ulem}
\usepackage{threeparttable}
\usepackage{tablefootnote}
\usepackage{color}
\usepackage{tabularx}
\usepackage{caption}

\def\bbr{{\Bbb{R}}} 

\def\bbs{{\Bbb{S}}}

\newcommand{\half}{ \mbox{\small$\frac{1}{2}$}}

\newcommand{\be}{\begin{equation}}
\newcommand{\ee}{\end{equation}}

\newcommand{\ccr}{{\mathfrak r}}

\def\e{\varepsilon}
\def\O{\Omega}

\def\tr {{\rm tr}}
\def\rank{{\rm rank}}
\def\dim {{\rm dim}}

\def\lin {{\rm lin}}

\newcommand{\V}{{\cal V}}

\newcommand{\C}{{\cal C}}
\newcommand{\A}{{\cal A}}

\newcommand{\G}{{\cal G}}
\newcommand{\J}{{\cal J}}
\newcommand{\M}{{\cal M}}

\newcommand{\LL}{{\cal L}}

\newcommand{\W}{\mathcal{W}}

\newcommand{\cM}{{\mathfrak M}}

\newcommand{\si}{{\mathfrak i}}

\def\bbc{{\Bbb{C}}}

\newcommand{\cL}{{\cal L}}
\newcommand{\N}{{\cal N}}

\newcommand{\T}{{\cal T}}

\newtheorem{theorem}{Theorem}[section]

\newtheorem{proposition}{Proposition}[section]

\newtheorem{remark}{Remark}[section]
\newtheorem{definition}{Definition}[section]

\newtheorem{cor}[theorem]{Corollary}

\def\yao{\textcolor{black}}

\def\rui{\textcolor{black}}

\begin{document}
\title{Goodness-of-fit tests on manifolds}
\author{Alexander Shapiro, ~
        Yao~Xie, ~Rui Zhang
\thanks{Alexander Shapiro (e-mail: \url{ashapiro@isye.gatech.edu}), Yao Xie (e-mail: \url{yao.xie@isye.gatech.edu}) and Rui Zhang (e-mail: \url{ruizhang_ray@gatech.edu}) are with the H. Milton Stewart School of Industrial and Systems Engineering, Georgia Institute of Technology, Atlanta, GA.}
\thanks{Research of   Alexander Shapiro was partly supported by NSF grant 1633196. Research of Yao Xie was partially supported by  NSF grants CCF-1442635, DMS-1938106, DMS-1830210,
and an NSF CAREER Award CCF-1650913.}
}



\maketitle

\begin{abstract}
We develop a general theory for the goodness-of-fit test to non-linear models. In particular, we assume that the observations are noisy samples of a submanifold defined by a
\yao{sufficiently smooth non-linear map}. The observation noise is additive Gaussian. Our main result shows that the ``residual'' of the model fit, by solving a non-linear least-square problem, follows a (possibly noncentral) $\chi^2$ distribution. The parameters of the $\chi^2$ distribution are related to the model order and dimension of the problem. We further present a method to select the model orders sequentially. We demonstrate the broad application of the general theory in machine learning and signal processing, including determining the rank of low-rank (possibly complex-valued) matrices and tensors from noisy, partial, or indirect observations, determining the number of sources in signal demixing, and potential applications in determining the number of hidden nodes in neural networks. 
\end{abstract}

\IEEEpeerreviewmaketitle

\section{Introduction}

Testing for goodness-of-fit of a model is a fundamental problem in statistics and signal processing (see, e.g., a survey in \cite{ding2018model}). The goal is to describe how well the model fits a set of observations. The model can be represented by a pre-specified distribution, or structured parametric models (such as time series or linear regression models). Commonly seen goodness-of-fit tests include the chi-square and Kolmogorov-Smirnov tests (see, e.g., \cite{Test05}). The goodness-of-fit test is often used for model diagnosis to determine the appropriate parsimonious models, for instance, selecting the order and type of time series models \cite{BrockwellDavis2010}. For linear regression, a related problem is variable selection \cite{hastie2005elements}, which determines a subset of variables that lead to the best overall fit to the data.

Although much has been done for model selection in linear models, it is unclear how to select  models given noisy observations in the non-linear setting, especially when there are underlying manifold structures. Such problems arise very often in machine learning and signal processing applications. For instance, how to select the rank of a low-rank matrix, decide the number of hidden nodes in neural networks, and determine the number of signal sources when observing their mixture.

In this paper, we develop a general theory for testing the goodness-of-fit of non-linear models. In particular, we assume that the observations are noisy samples of a submanifold (defined by a {\color{black} sufficiently smooth} non-linear map). The observation noise is additive Gaussian. Our main result shows that the ``residual'' of the model fit (by solving a non-linear least-square problem) follows a (possibly non-central) $\chi^2$ distribution. The parameters of the $\chi^2$ distribution are related to the model order and dimensions of the problem. 
 \yao{A key component of our analysis is the characteristic rank of the Jacobian matrix associated with the non-linear map that defines the submanifold. }
A natural use of our result is to the select order of a model via a sequential test procedure \yao{by choosing between two nested models}. We are particularly interested in ``nested'' models, i.e., one can order the models by their complexity. We demonstrate the applications of this general theory in the settings of real and complex matrix completion from incomplete and noisy observations, signal source identification, and determining the number of hidden nodes in neural networks.

It is worthwhile pointing out that the model goodness-of-fit test here is not the same as the widely known model order selection based on the celebrate AIC and BIC rules, etc.; the related field of is extensive (see, e.g., a recent survey in \cite{ding2018model}). The criterion for model order selection therein is the ``prediction'' or ``generalization'' error. In contrast, the goodness-of-fit we consider here is to describe how well a model fits a set of observations (thus, we consider ``residual'' errors). One potential issue with the classic model order selection based on AIC/BIC is that for certain situations, the expected prediction/generalization error may not be easily derived (for linear regression, there are explicit expressions). Such situations happen, for instance, when observations are noisy samples on a manifold. In such cases, the classic AIC and BIC rule may not be easy to carry through and may require significant numerical simulation to estimate the prediction errors. One benefit of the proposed approach is that the distribution of the residual is explicitly characterized. Thus, we can use it conveniently for selecting model orders through a sequential test procedure.

\yao{The proposed framework differs from other goodness-of-fit tests, such as the classic Kolmogorov-Smirnov test, which determines whether the empirical distribution is close a ``nominal'' or ``target'' distribution, the non-parametric approach based on the Maximum Mean Discrepancy (MMD) divergence \cite{chwialkowski2016kernel, liu2016kernelized, jitkrittum2017linear}, and the Bayesian approach \cite{verdinelli1998bayesian}. Our proposed framework also differs from the work on testing the manifold hypothesis  \cite{fefferman2016testing}, i.e., determining whether data lie near a low-dimensional manifold; \cite{fefferman2016testing} uses a ``worst-case'' analysis without assuming prior information about data generation mechanism.}

\yao{Part of our work is related to low-rank matrix completion from partial and noisy observations. There has been much work done in this field, with notable contributions of \cite{candes2010power, candes2010matrix, recht2010guaranteed,klopp2014noisy, DavenportRomberg16, pimentel2016characterization}. There are mainly two categories of algorithms, including convex relaxation based on nuclear norm minimization, and non-convex optimization based on alternating minimization. In nuclear minimization (see, e.g., \cite{candes2010matrix}), the rank selection is not explicitly addressed, possibly due to that the focus is on the recovery of the matrix itself. It is not clear how true rank will be recovered using the nuclear norm minimization approach. However, it is known that nuclear norm minimization may be asymptotically biased (see, e.g., \cite{sxz19}). For non-convex optimization-based matrix completion, such as alternating minimization \cite{hastie2015matrix}, one has to \rui{pre-determine} the rank of the matrix, and it is typically done empirically by heuristic methods \cite{recht2013parallel}.}

\textcolor{black}{Our proposed framework also differs from the work on testing the manifold hypothesis in \cite{fefferman2016testing}. The approach in \cite{fefferman2016testing} is nonasymptotic and,  in a sense, nonparametric. It is assumed there that the data is generated from a ``true" but unknown distribution. The algorithmic question addressed in \cite{fefferman2016testing} is, given a sample of size $N$, whether it is possible to verify with a high probability existence of a manifold, from a family of $d$-dimensional $\C^2$-submanifolds,  which fits the data with a prescribed accuracy measured in terms of an average squared distance. Unlike our approach, no parametric model is assumed, while a ``worst-case'' analysis is applied in \cite{fefferman2016testing}. On the other hand we consider a nested family of parametrically defined  manifolds.}

The rest of the paper is organized as follows. Section \ref{sec-gen} presents the background knowledge. Section \ref{sec-test} contains the main results: the test statistics for model selection on manifolds. \yao{Section \ref{sec-appl} gives several examples to demonstrate the use the general theory in specific settings. Section \ref{sec:eg} presents numerical experiments.} Finally, section \ref{sec:conclusion} concludes the paper with discussions on future directions.

Our notations are conventional. By $\|x\|_2$ we denote the Euclidean norm of vector $x\in \bbr^m$. By $\lin (A)$ we denote the linear space generated by columns of the matrix $A$ and by $\tr(A)$ the trace of the square matrix $A$. For a linear space $\LL\subset \bbr^m$, we denote by $\LL^\perp=\{y\in \bbr^m:y^\top x=0,\;x\in \LL\}$ its orthogonal space. All proofs are delegated to the Appendix.

\setcounter{equation}{0}
\section{Background}\label{sec-gen}

In this section, we present the general theory, which, in particular, will help to develop subsequent test statistics for determining model orders in Section \ref{sec-test}.

Consider  a nonempty   set $\Theta\subseteq\bbr^d$ and  a mapping $G:\Theta\to \bbr^m$. {\color{black} We assume throughout the paper that the set $\Theta$ is  {\em open  and connected}.}
Here, $d$ is the dimension of the parameter space (also referred to as the intrinsic dimension), and $m$ is the dimension of the observation space. {\color{black} Consider   a point $\hat{y}\in \bbr^m$} and  the least squares problem:
\begin{equation}\label{ls-1}
 \min_{\theta\in \Theta} \|\hat{y}-G(\theta)\|^2_2.
\end{equation}
Define the  image  of  the  mapping $G$,
\begin{equation}\label{image}
\cM:=\{G(\theta):\theta\in \Theta\}.
\end{equation}
Then problem \eqref{ls-1} can be written as
\begin{equation}\label{ls-2}
 \min_{x\in \cM} \|\hat{y}-x\|^2_2.
\end{equation}
That is, in problem \eqref{ls-2}, we aim to find a point of the set $\cM$ such that the Euclidean distance is minimized.
{\color{black} We deal with situations where the set $\cM$ is a {\em smooth manifold}; we will discuss this below.
By saying that the manifold is {\em smooth} we mean that it is at least $\C^2$ smooth.
}

We assume that the map  $G(\cdot)$ is  {\color{black} at least $\C^2$   smooth,  i.e., $G(\cdot)=(g_1(\cdot), \ldots, g_m(\cdot))$ with functions  $g_i:\Theta\to\bbr$, $i=1, \ldots ,m$,  being twice continuously differentiable.   In some cases we make the stronger assumption  that $G(\cdot)$ is
{\em analytic},   i.e., every $g_i(\cdot)$, $i=1, \ldots ,m$,  is analytic.}
Recall that a function is analytic on an open subset of $\bbr^d$, if it can be expanded in power series in a neighborhood of every point of this set. {\color{black} For instance,} every polynomial function is analytic.

With the mapping $G(\theta)$ is  associated the $m\times d$ Jacobian matrix
\begin{equation}\label{jacob}
 J (\theta):=\partial G(\theta)/\partial \theta,
\end{equation}
whose components are formed by partial derivatives
\[[J (\theta)]_{ij}=\partial g_i(\theta)/\partial \theta_j, ~i=1, \ldots, m, ~ j=1, \ldots, d.\] The differential of $G(\cdot)$ at a point $\theta\in \Theta$ is the linear mapping $dG(\theta):\bbr^d\to\bbr^m$ given by $dG(\theta) h=J(\theta)h$.

\begin{remark}
\label{rem-1}
{\rm
It is possible to deal with more general settings where the set $\Theta$ is a smooth connected manifold (without boundaries) rather than an open set. In that case, the derivations below can be pushed through by considering the corresponding Jacobian matrices in the local systems of coordinates of $\Theta$.}
\end{remark}

\begin{definition}[Characteristic rank]
\label{def-crank}
We refer to the maximal rank of the Jacobian matrix,
\begin{equation}\label{ls-3}
\ccr:=\max_{\theta\in\Theta} \{\rank (J(\theta))\},
\end{equation}
as the {\em characteristic rank} of the mapping $G(\cdot)$.
\end{definition}

{\color{black} The following Proposition \ref{pr-1} shows that, when $G(\cdot)$ is {\em analytic}, the characteristic rank in a certain sense is generic.}
By saying that a property holds for almost every (a.e.) $\theta\in \Theta$, we mean that there is a set $\Upsilon \subset \Theta$ of Lebesgue measure zero such that the property holds for all $\theta\in \Theta\setminus \Upsilon$.
Discussions of the following result can be found in \cite{sha1986}; we give its proof in the Appendix.

{\color{black}
\begin{proposition}
\label {pr-1}
The following holds: {\rm (i)}  The set
$\left \{\theta\in \Theta: \rank (J(\theta))=\ccr\right\}$ is open.
 {\rm (ii)} If  the map $G(\cdot)$ is {\rm analytic}, then
for a.e. $\theta\in \Theta$   the rank of the Jacobian matrix  $J(\theta)$ is equal to  the  characteristic rank  $\ccr$.
\end{proposition}

If $\rank(J(\theta_0))=\ccr$ for some $\theta_0\in \Theta$, then   there is a neighborhood of $\theta_0$ such that  $\rank(J(\theta))=\ccr$ for all $\theta$ in that neighborhood. It follows by the Constant Rank Theorem  (e.g., \cite{stern})   that there is  a neighborhood $\V$  of $\theta_0$ such that  the set $G(\V)$ forms a smooth  manifold of dimension $\ccr$,  in the space  $\bbr^m$,
with the tangent space generated by the columns of the Jacobian matrix $J(\theta)$.   When  the map $G(\cdot)$ is analytic,   if we choose a point $\theta_0$ at random, with respect to a continuous distribution on the set $\Theta$, then $\rank(J(\theta_0))=\ccr$ almost surely (with probability one).

\begin{remark}
\label{rem-sard}
{\rm Assuming that the mapping $G(\cdot)$ is $\C^\infty$ smooth,
we have  by \rui{Sard's theorem \cite{sards1942measure}} that   the image $\cM$  (of $G$)  has Lebesgure measure zero in the  observation space $\bbr^m$  if and only if  $\ccr<m$.}
\end{remark}
}

\begin{definition} [Regularity \cite{sha1986}]
\label{def-1}
We say that a point $\theta_0\in \Theta$  is   {\rm regular} if   rank of the Jacobian matrix $J(\theta_0)$ is equal to the characteristic rank $\ccr$ and moreover there exist  neighborhoods $\V$ of $\theta_0$ and $\W$ of $G(\theta_0)$  such that $\cM\cap \W=G(\V)$.
\end{definition}

The regularity of $\theta_0$ ensures that the local structure
of $\cM$ near $x_0=G(\theta_0)$ is provided by the mapping $G(\cdot)$ defined in a neighborhood of $\theta_0$. Hence, $\cM$ is a smooth manifold of the dimension of the characteristic rank $\ccr$, in a neighborhood of $x_0$.
In particular, this implies that if $\theta'\in \Theta$ is such that $G(\theta')=G(\theta_0)$, then there are neighborhoods $\V'$ of $\theta'$ and $\V_0$ of $\theta_0$ such that $G(\V')=G(\V_0)$.
 A result deeper than the one of Proposition \ref{pr-1}(ii)
 says that when the coordinate mappings $g_i(\cdot)$, $i=1, \ldots, m$, are analytic \yao{(for instance polynomial)} and either the set $\Theta$ is bounded or $G(\theta)\to \infty$ as $\theta\to \infty$, then a.e. point
$\theta_0\in \Theta$ is regular (e.g., \cite[Section 3.4]{fed69}).

We denote by $\T_\cM(x)$ the {\it tangent space} to $\cM$ at a point $x\in \cM$, provided $\cM$ is a smooth manifold
in a neighborhood of $x$.
Let $\theta_0$ be a regular point of $G(\cdot)$ and $x_0=G(\theta_0)$. Then $\T_\cM(x_0)=\lin (J(\theta_0))$ and dimension of $\T_\cM(x_0)$ is equal to the rank $\ccr$ of
$J(\theta_0)$.
Also,
$\T_\cM(x_0)$ coincides with the image of the differential $dG(\theta_0)$, i.e.,
\begin{equation}\label{tansp}
\T_\cM(x_0)=\left\{dG(\theta_0) h: h\in \bbr^d\right\}.
\end{equation}

\section{Test statistics on manifold}
\label{sec-test}

We view now the mapping $G(\theta)$ as a considered model of the parameter vector $\theta\in \Theta$, and problem \eqref{ls-1} as the least squares estimation (LSE) procedure with $\hat{y}$ being a given data point. More specifically, we assume the following model
\begin{equation}\label{mod-1}
 \hat{y}=x_0+N^{-1/2}\gamma +\e,
\end{equation}
where $x_0\in \cM$
is viewed as the population (true) value, vector
$\gamma\in \bbr^m$ is a deterministic bias, and the error vector $\e$ is random. When $\hat{y}$ is estimated from a random sample, the parameter $N$ represents the sample size. In general, $N$ can be viewed as a normalization parameter allowing to formulate rigorous convergence results for $N$ tending to infinity. We assume that the components $\e_i$, $i=1, \ldots, m$, of $\e$ are  independent of each other and such that $N^{1/2} \e_i$ converges in distribution, as $N\to\infty$, to normal distribution
with mean zero and variance $\sigma^2>0$.  The term
$N^{-1/2}\gamma$ represents systematic deviations form the ``true" model and is referred to in statistics literature as the population drift (e.g.,\cite{McManus}).

We consider the following least squares test statistic to determine the model
\begin{equation}\label{mod-2}
T_N:=N\hat{\sigma}^{-2}\min_{x\in \cM}\|\hat{y}-x\|_2^2,
\end{equation}
where $\hat{\sigma}^{2}$  is a consistent estimate of $\sigma^2$.

 \subsection{Test statistic on manifolds}

We now consider the general case defined in (\ref{mod-1}). We will show that for the problem defined on smooth manifolds, similar results in the form of RSS for linear models hold.

\begin{remark}
\label{rem-con}
{\rm
 For any $\hat{y}\in \bbr^m$, the generalized least-square problem \eqref{ls-2} has an optimal solution which may be not unique.
If $y_k$ is a sequence converging to $x_0\in \cM$ and $x_k$ is an optimal solution of \eqref{ls-2}, then $x_k$ converges to $x_0$ (e.g., \cite[Theorem 7.23]{SDR2014}).
Under the model \eqref{mod-1} we have that $\hat{y}$ converges to $x_0$ in probability as $N\to\infty$. It follows that any minimizer $\hat{x}$ in the right hand side of \eqref{mod-2} converges in probability to $x_0$.
}
\end{remark}

Suppose that $\cM$ is a smooth manifold in a neighborhood $\W$ of the point $x_0$. If $\hat{x}\in \W$ is an optimal solution of the least squares problem \eqref{mod-2}, then it follows that
\begin{equation}\label{nescon}
 \hat{y}-\hat{x}\in [\T_\cM(\hat{x})]^\perp,
\end{equation}
where $\T_\cM(\hat{x})$ is the respective tangent space (see (\ref{tansp})). The following result shows that  for $\hat{y}$ sufficiently close to $x_0$, the necessary optimality condition \eqref{nescon} is also sufficient
(cf., \cite[Proposition III.4]{sxz19}).

\begin{proposition}
\label{pr-uniq}
Suppose  that $\cM$ is a smooth   manifold  in a neighborhood of  $x_0\in \cM$. Then there exists a neighborhood  $\W$  of $x_0$ such that if $\hat{y}\in \W $ and  a point $\hat{x}\in \W\cap \cM$ satisfies condition \eqref{nescon}, then $\hat{x}$ is the unique globally optimal solution of the least squares estimation problem \eqref{mod-2}.
\end{proposition}

Since the least-squares problem in \eqref{mod-2} is non-convex,  standard optimization algorithms are at most guaranteed to converge to a stationary point satisfying first-order optimality conditions of the form \eqref{nescon}. The above proposition shows that if the fit is ``sufficiently good", then, in fact, the computed stationary point is globally optimal. Of course, this result is of a local nature, and it would be difficult to quantify what fit is good enough. Nevertheless, this tries to explain an empirical observation that for good fits, the problem of {\em local } optima does not happen too often.

Under the model \eqref{mod-1}
we have  the following asymptotic results, which are counterparts of the properties when $\cM$ is a linear space (cf.,  \cite{sha1986}).

\begin{theorem}[Asymptotic distribution of test statistic]
\label{th-asym}
Suppose  that $\cM$ is a smooth  manifold, of dimension $\ccr$,  in a neighborhood    of the point $x_0\in \cM$. Let $P$ be the orthogonal projection matrix  onto the tangent space $\T_\cM(x_0)$.
 Then the following holds as  $N\to\infty$:
  \begin{itemize}
  \item[\rm (i)] With probability tending to one  the least squares problem \eqref{mod-2}
has unique optimal solution $\hat{x}$,
\item[\rm (ii)]
 The test statistic $T_N$ in \eqref{mod-2} converges in distribution to the noncentral $\chi^2$ distribution with $m-\ccr$  degrees-of-freedom and the noncentrality parameter
$ \delta=\sigma^{-2}\| (I_m-P)\gamma\|_2^2$.
  \item[\rm (iii)] The scaled estimator
$N^{1/2}(\hat{x}-x_0)$ converges in distribution to a multivariate normal distribution with  the mean vector $P\gamma$ and the covariance matrix $\sigma^2 P$.
\item[\rm (iv)] The scaled error $N^{1/2}e$ converges in distribution to a multivariate normal distribution with the mean vector $(I_m-P)\gamma$ and the covariance matrix $\sigma^2 (I_m-P)$, where $e=\hat{y}-\hat{x}$ is a vector of residuals.
\end{itemize}
\end{theorem}

\subsection{Nested models}

Consider now nested models,  meaning the setting such that models can be naturally ordered by their complexity. For instance, the linear regression problems, one can sequentially increase or remove the variables being used in the model. Mathematically, this poses a natural order for the parameter space.  That is, let  $\Theta'\subset \Theta$ be a smooth manifold of dimension $d'$, and let
\[\cM':=\{ G(\theta):\theta\in \Theta'\}.\] Let $\theta_0\in \Theta'$  be a regular point of the mapping $G$. Then $\cM$ is a smooth manifold in a neighborhood of the point $x_0=G(\theta_0)$. Moreover, $\cM'$ forms a smooth submanifold in a neighborhood of the point $x_0$ with the tangent space (compare with \eqref{tansp})
\begin{equation}\label{mod-4}
 \T_{\cM'}(x_0)=\left\{dG(\theta_0)h:h\in \T_{\Theta'}(\theta_0)\right\}.
\end{equation}
Note that $ \T_{\cM'}(x_0)\subseteq \T_{\cM}(x_0)$ and
it could happen that $ \T_{\cM'}(x_0)= \T_{\cM}(x_0)$ even  when $d'<d$.

Consider now the test statistic
\begin{equation}\label{mod-5}
T'_N:=N\sigma^{-2}\min_{x\in \cM'}\|\hat{y}-x\|_2^2.
\end{equation}
We have  the following results  (cf., \cite{ste1985}).

\begin{theorem}
\label{th-nest}
Suppose that $\cM$ is a smooth  manifold of dimension $\ccr$
and $\cM'\subset \cM$ is a smooth  manifold of dimension $\ccr'$,
 in a neighborhood of the point $x_0\in \cM'$.
 Then the following holds:
 \begin{itemize}
 \item[\rm (i)] $T'_N$ converges in distribution to a noncentral $\chi^2$ random variable with $m-\ccr'$  degrees-of-freedom and the noncentrality parameter $\delta'= \sigma^{-2}\| (I_m-P')\gamma\|_2^2$, where $P'$ is  the orthogonal projection matrix  onto the tangent space $\T_{\cM'}(x_0)$.
 \item[\rm (ii)] The difference statistic
   $T'_N-T_N$ converges in distribution to a noncentral $\chi^2$ random variable with $(m-\ccr')-(m-\ccr)=\ccr-\ccr'$  degrees-of-freedom and the noncentrality parameter $\delta'-\delta$.
 \item[\rm (iii)] The statistics $T'_N-T_N$ and $T_N$ are asymptotically independent.
 \end{itemize}
  \end{theorem}

\subsection{Decomposable maps}
\label{sec-spec}
Now we will make additional structural assumptions about the mapping that defines the manifold of our problem. We will make sense of such structural decompositions in specific applications in Section \ref{sec-appl}.
Consider model defined by the following mapping
\begin{equation}\label{sp-1}
G(\theta):=\G(\xi)+\A(\zeta),
\end{equation}
where $\Xi\subseteq\bbr^d$ is a nonempty open connected  set, $\G:\Xi\to \bbr^m$ is a smooth mapping and $\A:\bbr^k\to \bbr^m$ is a linear mapping. {\color{black} Note that $G(\cdot)$ inherits smoothness properties of $\G(\cdot)$. In particular, if $\G(\cdot)$ is analytic, then the corresponding mapping  $G(\cdot)$  is analytic.}

The parameter vector here is  $\theta=(\xi,\zeta)$ and the parameter space $\Theta =\Xi\times \bbr^k$. We assume that $\A(\zeta)=A\zeta$, where $A$ is an $m\times k$ matrix of rank $k$. Denote by
$$\M:=\{\G(\xi):\xi\in \Xi\}\; {\rm and} \; \LL:=\{\A(\zeta):\zeta\in \bbr^k\}$$
  the images of the mappings $\G$  and $\A$, respectively.
 Note that the linear space $\LL$ has a dimension $k$, and $\cM=\M+\LL$ is the image of the mapping $G:\Theta\to \bbr^m$. We denote by $\ccr$ the characteristic rank of mapping $G(\cdot)$, and
 by $\rho$ the {\it characteristic rank} of  $\G(\cdot)$, i.e.,
 \begin{equation}\label{rankrho}
\rho:=\max_{\xi\in \Xi}\rank(\J(\xi)),
 \end{equation}
 where $\J(\xi)=\partial \G(\xi)/\partial \xi$  is   the Jacobian of $\G(\cdot)$.

Consider the corresponding least squares problem \eqref{ls-2},  the model \eqref{mod-1} and the least squares test statistic $T_N$, defined in  \eqref{mod-2}, for the mapping $G(\theta)$ of the form \eqref{sp-1}.
\begin{remark}
\label{rem-2}
{\rm
Note that the optimal value of least squares problem \eqref{ls-2} is not changed if the point $\hat{y}$ is replaced by $\hat{y}+v$ for any $v\in \LL$. Therefore the corresponding  test statistic $T_N$ can be considered as a function of $\hat{y}'=P_{\LL^\perp}\hat{y}$, where $P_{\LL^\perp}=I_m-P_{\LL}$ is the orthogonal projection onto the linear space orthogonal to $\LL$.
}
\end{remark}

Recall that $\cM=\M+\LL$. If $\cM$ is a smooth manifold, of dimension $\ccr$, in a neighborhood of $x_0$,  then  Theorems \ref{th-asym} and \ref{th-nest} can be applied. In particular, it will follow that the test statistic $T_N$ converges in distribution to a noncentral $\chi^2$ with $m-\ccr$  degrees-of-freedom and certain noncentrality parameter.

Note that  for $\theta=(\xi,\zeta)\in \Theta$,  the differential $dG(\theta):\bbr^d\times \bbr^k\to\bbr^m $ is given by
 \begin{equation}\label{diff}
 dG(\theta)(h,z)=d\G(\xi)h+A z,\;h\in \bbr^d, z\in \bbr^k.
 \end{equation}
This implies  that the corresponding characteristic rank
 $\ccr\le \rho+k $.

  \begin{definition}
\label{def-wellp}
We say that a point $x\in \M$  is {\em well-posed} if $\M$ is a smooth manifold of dimension $\rho$ in a neighborhood of $x$ and
\begin{equation}\label{cond}
\T_\M(x)\cap \LL=\{0\}.
\end{equation}
We say that the model is {\em well-posed} if
\begin{equation}\label{well}
\ccr=  \rho+k.
\end{equation}
\end{definition}
For the matrix completion problem the well-posedness condition (at a point)  was introduced in \cite{sxz19}.  Note that condition \eqref{cond} means that
\begin{equation}\label{dimcon}
 \dim(\T_\M(x)+ \LL)=\dim(\T_\M(x))+\dim(\LL).
\end{equation}
{\color{black}
Of course, a necessary condition for \eqref{dimcon} to hold is that 
$\rho+k\le m$. Note also that assuming the mapping $\G(\cdot)$, and hence the mapping $G(\cdot)$, is analytic we have that the image $\cM=\M+\LL$ has Lebesgue measure zero in the observation space $\bbr^m$ if and only if $\ccr<m$ (see Remark \ref{rem-sard}).
 }

\begin{proposition}
\label{pr-wpos}
{\color{black} Suppose that the mapping  $\G(\cdot)$  is analytic. Then the following holds.}
If there exists at least one well-posed point $x\in \M$,  then
the model is well-posed. Conversely if $\M$ is a smooth manifold of dimension $\rho$ and the model is well-posed, then for a.e. $\xi\in \Xi$, the corresponding point $x=\G(\xi)$ is well-posed.
 \end{proposition}

Let us make the following observation. By the definition of $\cM$ under the decomposition \eqref{sp-1}, we have that the point $x_0\in \cM$ can be represented as
\begin{equation}\label{sp-2}
x_0=x^*+v_0 \;\text{for some}\;x^*\in \M,v_0\in \LL.
\end{equation}

\begin{definition}
\label{def-id}
We say that the model  is {\em identifiable} at $x^*$ (at $x_0$)  if the representation \eqref{sp-2}  is unique, i.e., if
 $x_0=x'+v'$ with  $x'\in \M$ and $v'\in \LL$, then $x'=x^*$. We say that  the model is {\em locally  identifiable} at $x^*$, if such uniqueness holds locally, i.e., there is a neighborhood $\W$ of $x^*$ such that  if  $x_0=x'+v'$ with   $x'\in \M\cap \W$ and $v'\in \LL$, then $x'=x^*$.
\end{definition}

The following result can be proved in the same way as \cite[Theorem III.2]{sxz19}.
\begin{proposition}
\label{pr-locid}
If a point $x^*\in \M$ is well-posed, then  the model is  locally  identifiable at $x^*$.
\end{proposition}

{\color{black} To verify the (global) identifiability of a nonlinear model is difficult, and often is out of reach. Of course,  local identifiability is a necessary condition for global  identifiability. When $\G(\cdot)$  is analytic, the well-posedness condition \eqref{well}  can be verified numerically;    it is necessary and sufficient for the local identifiability in the  generic sense of Proposition \ref{pr-wpos}. We argue that  the well-posedness condition  is a minimal property  that should be verified for a considered model.
}

\section{Applications of general theory}
\label{sec-appl}

In this section, we present several examples in signal processing and machine learning to illustrate how to use the general theory, developed in the previous section, to determine the ``model order'' in the specific setting.

\begin{remark}
\label{rem-rank}
{\rm
For some well-structured manifolds, it is possible to give an explicit formula for the characteristic rank.
In more complicated settings, we can find the characteristic rank numerically. That is, we compute the Jacobian matrix of the considered mapping at several randomly generated points of $\Theta$, and subsequently compute its rank. By Proposition \ref{pr-1},
we can expect that this will give us the characteristic rank of the considered mapping. This approach worked quite well in experiments reported in Sections \ref{complex}, 
 \ref{mat-sen},
and \ref{seismic} below.
}
\end{remark}


\subsection{Noisy matrix completion}
\label{mat-com}

We first show that the problem of selecting rank for noisy matrix completion can be addressed using our general theory. Part of the relevant discussion can be found in \cite{sxz19}; here, we generate a conclusion using the framework of our general theory in this paper.

Consider the noisy matrix completion problem
(e.g., \cite{candes2010power},
\cite{DavenportRomberg16},\cite{recht2010guaranteed}
and references there in).
Suppose we observe a subset of entries of a low-rank matrix with Gaussian noise and aim to recover the matrix. A common approach to solve this problem, is to use a matrix factorization by selecting a rank of the matrix using subjective choice or experiments and cross-validation. However, it is not clear what would be a good statistical procedure to determine the rank of the matrix.

Consider a mapping $G(\theta)$  of the form \eqref{sp-1} with the following parameters.
Let $\xi=(V, W)$ with $V\in \bbr^{n_1\times r}$ and $W\in  \bbr^{n_2\times r}$,  $r\le\min\{n_1,n_2\}$,
and let  $\Xi\subset \bbr^{n_1\times r}\times \bbr^{n_2\times r}$ be the set
of such $\xi$ with both matrices $V$ and $W$ having full column rank $r$.
Define
\[\G(\xi):=V W^\top\in \bbr^{n_1\times n_2},\] and
\[\LL:=\{X\in\bbr^{n_1\times n_2}: X_{ij}=0,\;(i,j)\in \O\},\]
for an index set $\O\subset \{1, \ldots ,n_1\}\times \{1, \ldots, n_2\}$.  Then $\M=\M_r$ forms the set of $n_1\times n_2$ matrices of rank $r$. Note that the set  $\Xi$ is an open   connected subset of $\bbr^{n_1\times r}\times \bbr^{n_2\times r}$, and
\[\dim(\LL)=n_1n_2-|\Omega|,\] where $|\Omega|$ is the cardinality  (number of elements) of the index set $\O$.
The parameter set \[\Theta=\Xi\times \bbr^{n_1n_2-|\Omega|}.\]

Here the least squares problem of \eqref{mod-2}, associated with the test statistic $T_N$, can be written as
\begin{equation}\label{mat-1}
\min_{X\in \M_r}\sum_{(i,j)\in \O}(\hat{Y}_{ij}-X_{ij})^2,
\end{equation}
where   $\hat{Y}_{ij}$, $(i,j)\in \O$, are observed values of the data matrix.
Then the model \eqref{mod-1} can be written as
\begin{equation}\label{mat-2}
\hat{Y}_{ij}=X^*_{ij}+N^{-1/2}\Gamma_{ij}+\e_{ij}, \;(i,j)\in \O,
\end{equation}
where $X^*\in \M_r$.  Note that  here  the  test statistic $T_N$ is a function of the components $\hat{Y}_{ij}$, $(i,j)\in \O$, of the corresponding matrix $\hat{Y}$ (compare with Remark \ref{rem-2} and \eqref{sp-2}).

It is well known that
the set $\M_r$, of $n_1\times n_2$ matrices of rank {\color{black} $r>0$,
 is a smooth manifold of dimension $r(n_1+n_2-r)$  in a neighborhood of its every point (excluding origin).}
Therefore here every  $\xi\in \Xi$ is a regular point of the  mapping $\G(\cdot)$ with the characteristic rank $\rho= r(n_1+n_2-r)$.
Thus for  the characteristic rank $\ccr$ of the corresponding mapping $G(\cdot)$ we have that
\begin{equation}\label{mat-3}
\ccr\le  r(n_1+n_2-r)+n_1n_2-|\Omega|,
\end{equation}
and that the model is well-posed if and only if the equality holds in \eqref{mat-3}.

 Let us make the following assumption.
 \begin{itemize}
   \item [(A)]
The set  $\cM=\M_r+\LL$ is a smooth manifold, of dimension $\ccr$,
 in a neighborhood of  the point $X$.
 \end{itemize}
 Note that if \yao{Assumption (A) holds,}
  then  $\cM $ is a smooth manifold  of dimension $\ccr$ in a neighborhood of $X'=X+U$ for any $U\in \LL$. Therefore by the discussion of Section \ref{sec-gen},  the above assumption (A) holds generically.
By  Theorem \ref{th-asym}
we have the following result as $N$ tends to infinity  (cf.,  \cite{sxz19}).
\begin{proposition}
  Suppose that Assumption (A) holds. Then
  the test statistic $T_N$ converges in distribution to a noncentral $\chi^2$ with degrees-of-freedom $n_1n_2-\ccr$
and the noncentrality parameter 	
\begin{equation}\label{nonpar}
\delta=\sigma^{-2}\min_{H\in\T_{ \M_r(X^*)}}\sum_{(i,j)\in \O}(\Gamma_{ij}-H_{ij})^2.
\end{equation}
\end{proposition}

Moreover, applying Proposition \ref{pr-wpos}, we can conclude the following under the assumption:
 \begin{itemize}
   \item [(B)]
 The point $X^*$ is well-posed and the model is identifiable at $X^*$.
 \end{itemize}

\begin{proposition}
Suppose  that Assumption (B) holds.
Then: (i) the equality holds in \eqref{mat-3}, (ii) the test statistic $T_N$ converges in distribution to noncentral $\chi^2$ with degrees-of-freedom  $|\Omega|-r(n_1+n_2-r)$
and the noncentrality parameter $\delta$ given in \eqref{nonpar},
(iii) with probability tending to one, problem \eqref{mat-1} has a unique optimal solution $\{\hat{X}_{ij}\}_{(i,j)\in \O}$.
\end{proposition}

The difference test statistic can be applied to the following setting. Consider another index set
$\O'\subset \{1, \ldots, n_1\}\times \{1, \ldots, n_2\}$ of cardinality $|\Omega'|$ such that $\O\subset \O'$ and the corresponding space
\[\LL':=\{X\in\bbr^{n_1\times n_2}: X_{ij}=0,\;(i,j)\in \O'\}.\] Clearly, $\LL'$ is a subspace of $\LL$, and the corresponding set
\[\Theta' =\Xi\times \bbr^{n_1n_2-|\Omega'|},\]  is a linear subspace of the set $\Theta$. By Theorem \ref{th-nest} we have the following.
\begin{proposition}\label{pr-matrix}
Suppose  that  Assumption (A) holds and moreover
  $\cM'$ is a smooth   manifold, of dimension $\ccr'$,
 in a neighborhood of  $X^*\in \cM'$.
Then  the difference   statistic
   $T'_N-T_N$ converges in distribution to noncentral $\chi^2$ with   degrees-of-freedom $\ccr-\ccr'$ and the  noncentrality parameter $\delta'-\delta$. Moreover,
the statistics $T'_N-T_N$ and $T_N$ are asymptotically independent.
\end{proposition}
 The above result can be used to compare the goodness-of-fit of two models.

{\rui{
\begin{remark}
{\rm
An application of Theorem \ref{th-nest} and Proposition \ref{pr-matrix} allows to estimate $\sigma^2$ when the variance of the noise is unknown. Specifically, let's assume $N=1$, $\Gamma_{ij} = 0$ and $\varepsilon_{ij}$ follows normal distribution with zero mean and variance $\sigma^2$. Denote the set of observation indices as $\Omega'$. By leaving out some observation, we have a new set of observation indices $\Omega$ such that $\Omega\subset \Omega'$. Then we can construct the estimate of $\sigma^2$ as the following: 
 \begin{align*}
   \tilde T_N' =& \min_{X\in\mathcal M_r} \sum_{(i,j)\in\Omega'}
    (\hat Y_{ij} - X_{ij})^2,\nonumber\\
    \tilde T_N =& \min_{X\in\mathcal M_r} \sum_{(i,j)\in\Omega}
    (\hat Y_{ij} - X_{ij})^2,\nonumber\\
     \hat\sigma^2 =& \frac{\tilde T_N' - \tilde T_N}{|\Omega'| - |\Omega|}.
 \end{align*}
 By Theorem \ref{th-nest} and Proposition \ref{pr-matrix}, we have $\sigma^{-2}(\tilde T'_N-\tilde T_N)$ follows a  $\chi^2$ distribution with degrees-of-freedom $|\Omega'| - |\Omega|$ asymptotically for the true model. Therefore, $\hat\sigma^2$ is a consistent estimator of $\sigma^2$. This method can be generalized to the other applications in this paper and more discussion is provided in the Appendix.}
\end{remark}
}

\subsection{Complex noisy matrix completion}
\label{complex}

In this section, we {\color{black} generalize the results} to ``complex matrix completion.'' Here, the observations and underlying low-rank matrices are over the field $\bbc$  of complex numbers.
 %
Consider the matrix completion problem (over complex numbers), where $X\in \bbc^{n_1\times n_2}$, $V\in \bbc^{n_1\times r}$, $W\in \bbc^{n_2\times r}$:
 \begin{equation}\label{gen3}
 \min_{V, W} \|X-V W^\top\|^2_2\;\;{\rm s.t.}\;X_{ij}=b_{ij},\;(i,j)\in \O.
\end{equation}
This can be formulated in terms of a real numbers problem as follows. Write
 \[V=V_1+\si V_2,\] where $\si^2=-1$,
 $V_1\in \bbr^{n_1\times r}$, and $V_2\in \bbr^{n_1\times r}$ are the real and imaginary parts of matrix $V\in \bbc^{n_1\times r}$. Similarly, let
 \[W=W_1+\si W_2, \quad X=X_1+\si X_2.\] Then $$V W^\top=(V_1W_1^\top-V_2 W_2^\top)+\si (V_1 W_2^\top +V_2 W_1^\top).$$
 Define
 \[\cL_1 := \{U\in \bbr^{n_1\times n_2}: U_{ij}=0, (i,j)\in\O\},\] and $\cL = \cL_1\times \cL_1$. Then we can set   \[\theta=(V_1,W_1,V_2,W_2, U_1,U_2),\] and  mapping $$G(\theta):= (G_1(\theta), G_2(\theta)),$$ where \[G_1(\theta) = V_1W_1^\top-V_2 W_2^\top + U_1,\] \[G_2(\theta)=V_1 W_2^\top +V_2 W_1^\top+U_2,\] and $U_1 \in \mathcal L_1$ and $U_2 \in \mathcal L_1$.
Hence we can write the problem \eqref{gen3} in the following form
\begin{equation}\label{gen4}
\begin{array}{cll}
 \min\limits_{\theta}&
   \|X_1-G_1(\theta)\|^2_2
   +  \|X_2-G_2(\theta)\|^2_2
  \\{\rm s.t.} &  X_{1,ij}=b_{1,ij}, X_{2,ij}=b_{2,ij},\;(i,j)\in \O,\\ &X_{1,ij} = X_{2,ij}=0,\;(i,j)\in\O^c.
   \end{array}
\end{equation}

The dimension of the  manifold of $n_1\times n_2$  complex matrices of rank $r$, in terms of real numbers, is twice the corresponding dimension $r(n_1+n_2-r)$ in the real case. That is, the characteristic rank  of the respective  mapping $\G(\cdot)$ here is
\[\ccr = 2r(n_1+n_2-r).\]  Note that this differs from the real-value matrix completion case in Section \ref{mat-com} by a factor of 2.

\subsection{Low-rank matrix sensing}
\label{mat-sen}

Matrix sensing problems \cite{li2017algorithmic} is related to matrix completion, where the observations are linear projections of the underlying low-rank matrix. Specifically, denote by $\bbs^{d\times d}$ the space of $d\times d$ symmetric matrices, and $\big<A, B\big>:=\tr(AB)$ the scalar product of $A,B\in \bbs^{d\times d}$.
 Let $X^*\in \bbs^{d\times d}$ be a  positive semidefinite matrix of rank $r$  needed to be recovered. Given  measurement matrices $A_i\in \bbs^{d\times d}$, $i=1, \ldots, m$,
we observe $y\in\bbr^m$, such that \[y_i=\langle A_i, X^*\rangle.\] Then we aim to solve the following least square problem.
\begin{equation}
	\label{matrix-sensing}
	\min_{U\in\bbr^{d\times r}}f(U):=\sum_{i=1}^m\left (y_i - \langle A_i, UU^\top \rangle \right)^2.
\end{equation}
It is shown in \cite{li2017algorithmic} that \eqref{matrix-sensing} is the same problem as the problem of fitting one-layer neural networks with quadratic activation in \eqref{ohl-ls}, which we discuss next.

\subsection{One-hidden-layer neural networks}
\label{neural}

We will show the general theory can be applied to determine the number of hidden nodes. Consider a one-layer neural networks. Let $x_i \in\bbr^d$ be the inputs and the observation is assume to be generated by:
\begin{equation}
\label{ohl}	
y_i = \mathbf 1^\top q(U^{*\top}x_i) +\varepsilon_i,
\end{equation}where $U^*\in\bbr^{d\times r}$, $\mathbf 1\in\bbr^r$ with all entries equal to 1 and $\varepsilon_i$ is the Gaussian noise with mean zero and variance $\sigma^2$. The activation function can be one of the following, \begin{enumerate}
	\item[(i)] Quadratic activation: \[q(z_1,\cdots,z_r) =(z_1^2, z_2^2,\cdots,z_r^2).\]
	\item[(ii)] Sigmoid activation: \[q(z_1,\cdots,z_r) =(1/(1+e^{-z_1}),\cdots, 1/(1+e^{-z_r})).\] 
\end{enumerate}
A commonly used approach to fit neural networks is to solve the least square problem:
\begin{equation}
\label{ohl-ls}		\min_{U\in\bbr^{d\times r}}f(U):=\sum_{i=1}^m\big(y_i - \mathbf 1^\top q(U^{\top}x_i)\big)^2.
\end{equation}
Define $\Theta = \bbr^{d\times r}$, for $U\in\Theta$,
\[G(U) = (g_1(U), \dots, g_m(U)),\] where $g_i(U) = \mathbf 1^\top q(U^{\top}x_i)$.
In this setting problem (\ref{ohl-ls}) becomes a least squares problem of the form   (\ref{ls-1}).

It is difficult to evaluate the characteristic rank $\ccr$ of the mapping $G$ in a theoretical way. By computing the rank of the corresponding Jacobian matrix (see Remark \ref{rem-rank}), we find the following formulas for the characteristic rank fit well in numerical experiments:
\[\ccr=dr - r(r-1)/2,\] for the Quadratic activation function; and
$\ccr=dr$ for the Sigmoid activation function. 

\subsection{Tensor completion}
\label{tensor}

 Next, we consider the problem of  determining  the rank of a tensor from incomplete and noisy observations to illustrate the role of the general theory.

Consider a tensor $X\in \bbr^{n_1\times \cdots\times n_d}$ of order $d$ over the field of real numbers. It is said that $X$ has rank one if
\[
X=a^1\circ\cdots \circ a^d,
\]
where $a^i\in \bbr^{n_i}$ is $n_i\times 1$ vector, $i=1, \ldots, d$, and $``\circ"$ denotes the vector outer product. That is, every element of tensor $X$ can be written as the product \[X_{i_1, \ldots ,i_d}= a^1_{i_1}\times \cdots\times a^d_{i_d}.\]
The smallest number $r$ such that tensor $X$ can be represented as a sum $X=\sum_{i=1}^r Y_i$
of rank one tensors $Y_i$ is called the rank of  $X$, and the corresponding decomposition is often referred to as the (tensor) rank decomposition, minimal CP decomposition, or Canonical Polyadic Decomposition (CPD).

The {\it tensor completion problem} can be formulated as the problem of reconstructing tensor of rank $r$ by observing a relatively small number of its entries.
The second order tensor (i.e., when  $d=2$) can be viewed as a matrix, and  this becomes the matrix completion problem discussed in Section \ref{mat-com}. Consider now third order tensors $X\in \bbr^{n_1\times n_2\times n_3}$, and denote by $\M_r$ third order tensors of rank $r$.
Without loss of generality, we can assume that $n_1\ge  n_2\ge n_3$.
With tensor $X\in \M_r$  are associated matrices $A\in \bbr^{n_1\times r}$, $B\in \bbr^{n_2\times r}$, $C\in \bbr^{n_3\times r}$ such that \[X=A\otimes B\otimes C,\] meaning that
\[
X=\sum_{i=1}^r a^i\circ b^i\circ c^i,
\]
with $a^i$, $b^i$, $c^i$ being $i$th columns of the respective  matrices $A$, $B$, $C$.

The above leads to the following parameterization of $\M_r$. For \[\xi=(A,B,C)\in \bbr^{n_1\times r}\times  \bbr^{n_2\times r}\times \bbr^{n_3\times r},\] consider mapping
\[\G(\xi):=A\otimes B\otimes C.\] By definition of the tensor rank we have that rank of tensor $X=\G(\xi)$ cannot be larger than $r$. So we define the parameter set
\begin{equation}\label{param}
\Xi:=\left \{\xi\in \bbr^{n_1\times r}\times  \bbr^{n_2\times r}\times \bbr^{n_3\times r}:\G(\xi)\in \M_r\right\}.
\end{equation}
 We need to verify that the set $\Xi$ is open and connected.
 Note that it could happen that the complement $(\bbr^{n_1\times r}\times  \bbr^{n_2\times r}\times \bbr^{n_3\times r})\setminus \Xi$ of the set $\Xi$, has positive (Lebesgue) measure, or even that $\Xi$ has measure zero.

Careful analysis of properties of  $\M_r$ is not trivial and is beyond the scope of this paper. We will make some comments below. Let us consider the following examples. Suppose that $n_3=1$. In that case, assuming that the elements of a matrix $C\in \bbr^{1\times r}$ are nonzero, by rescaling columns of the respective matrices $A$ and $B$, we can assume that all elements of $C$ equal 1. Consequently, essentially, this becomes the matrix completion problem discussed in Section \ref{mat-com}. Thus the characteristic rank of $\G(\xi)$ in that case is $\ccr= r(n_1+n_2-r)$.

The key question of the tensor rank decomposition is its uniqueness. Clearly the decomposition $X=A\otimes B\otimes C$, of $X\in \M_r$, is invariant with respect to permutations, and rescaling of the columns of matrices $A$, $B$, $C$ by factors $\lambda_{1i},\lambda_{2i},\lambda_{3i}$, $i=1, \ldots, r$, such that
$\lambda_{1i}\lambda_{2i}\lambda_{3i}=1$.
It is said that the decomposition $X=A\otimes B\otimes C$ is (globally) {\em identifiable} if it is unique up to the corresponding permutation and rescaling. It is beyond the scope of this paper to give a careful discussion of the (very nontrivial) problem of tensor rank identifiability. As it was pointed above, for $n_3=1$ this becomes the matrix rank problem for which the identifiability never holds for $r>1$ (e.g., \cite[section 3.2]{kolda2009tensor}).

Suppose now that $n_3\ge 2$.
In that case  the situation is different.
\begin{definition}
\label{def-tenuniq}
 It is said that the rank $r$ decomposition is {\em generically} identifiable if for almost every
  $(A,B,C)\in \bbr^{n_1\times r}\times   \bbr^{n_2\times r}\times \bbr^{n_3\times r}$  the corresponding tensor $A\otimes B\otimes C$ has identifiable rank $r$.
  \end{definition}
In particular, the generic identifiability implies that the complement of the parameter set $\Xi$, defined in \eqref{param}, has (Lebesgue) measure zero. It is known that for sufficiently small $r$, the identifiability holds in the generic sense (we refer to \cite{chian2017},\cite{doman}, and references therein for a discussion of the tensor rank identifiability from a generic point of view).

The identifiability is related to the characteristic rank:
\begin{definition}
\label{def-lociden}
We say that $(A,B,C)\in  \bbr^{n_1\times r}\times   \bbr^{n_2\times r}\times \bbr^{n_3\times r} $ is  {\em  locally  identifiable}
if there is a neighborhood $\W$ of $(A,B,C)$ such that
$(A',B',C')\in \W$ and $A'\otimes B'\otimes  C'=A\otimes B\otimes  C$ imply  that $(A',B',C')$ can be obtained from $(A,B,C)$ by the corresponding rescaling.
We say that model $(n_1,n_2,n_3,r)$   is  {\em generically} locally   identifiable if a.e. $(A,B,C)\in  \bbr^{n_1\times r}\times   \bbr^{n_2\times r}\times \bbr^{n_3\times r} $ is  locally  identifiable.
\end{definition}
Note that  local identifiability of
 $(A,B,C)\in  \bbr^{n_1\times r}\times   \bbr^{n_2\times r}\times \bbr^{n_3\times r} $ is a local property, it could happen
  that rank of the corresponding  tensor
   $A\otimes B\otimes C$ is less than  $r$. If indeed the rank of tensor $A\otimes B\otimes C$ is $r$, then its global identifiability implies its local identifiability (note that the permutation invariance  does not affect  the local identifiability).
Note also that  the rank of the Jacobian matrix of a  mapping $\G(\xi)$ is always less than or equal to $r(n_1+n_2+n_3)-2r$.  This follows by counting the number of elements in $(A,B,C)$ and making corrections for the scaling factors. That is, the characteristic rank $\ccr$  of  map   $\G(\cdot)$  cannot be larger than $r(n_1+n_2+n_3-2)$.

\begin{proposition}
\label{pr-char}
Model $(n_1,n_2,n_3,r)$  is  generically locally   identifiable if and only if the following formula for the  characteristic rank $\ccr$ holds,
\begin{equation}\label{charrank}
\ccr=r(n_1+n_2+n_3-2).
\end{equation}
\end{proposition}

Since the generic (global)  identifiability implies generic local identifiability we have  the following consequence of the above proposition.
\begin{cor}
If the rank $r$ decomposition is  generically  identifiable, then formula \eqref{charrank} for the characteristic rank  follows.
\end{cor}


\subsection{Determining number of sources in blind de-mixing problem}
\label{seismic}

De-mixing problem (e.g., \cite{7919265}) is a fundamental challenge in signal processing, which arises from applications such as ambient noise seismic imaging \cite{snieder2010imaging}, NMR imaging, etc. In such problems, the goal is to recover the signals by observing their weighted mixture. Blind de-mixing is particularly challenging in which we do not know the waveforms of the signal. Moreover, the number of signals and the magnitudes of the waveforms are also unknown. Such a problem has been addressed using a matrix factorization approach \cite{mccoy2014sharp}. However, in existing approaches, there is no efficient method to determine the number of signals, which is usually a critical input parameter to algorithms. In this section, we show how to determine the number of sources in the context of ambient noise imaging using the general theory.

Assume there are $N$ sensors. Define the signal received by the $n$th sensor as follows:
\begin{equation}
x_n(t) = \sum_{k=1}^K s_k(t - \tau_{n,k}), \quad n = 1, \ldots, N.
\label{model}
\end{equation}
Assume the number of signals $K$ and the delays $\tau_{n,k}$ are all unknown. Further assume the signal is a Gaussian function
\[s_k(t) = \rho_k e^{-\alpha_k t^2},\] where $\alpha_k$ defines the width of the $k$th source, and $\rho_k$ is the magnitude of the $k$th source. Here, our goal is to estimate the number of signal sources $K$ from observations of $x_n(t)$ buried in Gaussian noise.

We now derive the observation model. For the ease of presentation, we present the derivation in continuous time (and continuous frequency) domain, and the switch to discrete-time (and discrete frequency) domain later. Let the Fourier transform of the signal to be
\begin{eqnarray*}
	S_k(f) := \mathcal F\{s_k(t)\}(f) = \int^\infty_{-\infty}s_k(t)e^{-2\pi \si tf}dt.
\end{eqnarray*}
Recall that the Fourier transform of the delayed signal corresponds to a phase-shift. 
Hence, for Gaussian signals in (\ref{model}), it can be shown that \[\mathcal F\{s_k(t-\tau)\}(f) = \rho_k\sqrt{\frac{\pi}{\alpha_k}} e^{-2\pi \si f\tau} e^{-\pi^2f^2/\alpha_k}.\]
For continuous function $h_1$ and $h_2$, the cross-correlation is defined as:
\[
	(h_1\otimes h_2)(s) := \int_{-\infty}^\infty h_1(t-s)h_2(t)dt.
\]
Here, in this section, $\otimes$ represents the cross-correlation operator. By the duality of convolution in frequency and time, we have
\[
	\mathcal F\{h_1\otimes h_2\}(f) = \mathcal F\{h_1\}^*(f) \mathcal F\{h_2\}(f),\]
where $(\cdot)^*$ denotes the conjugate of a complex number.

In ambient noise imaging, the useful ``signal'' are extracted by performing pairwise cross-correlation between sensors. Define $r_{n,m}(t)$ as the cross-correlation function of the $n$th and the $m$th sensors:
\begin{eqnarray*}
\begin{split}
r_{n,m}(t) &= x_n(t)\otimes x_m(t) \\
& = \sum_{k=1}^K\sum_{l=1}^K s_l(t - \tau_{n,l})\otimes s_k(t - \tau_{m,k}).
\end{split}
\end{eqnarray*}
Now consider the frequency domain. Denote the Fourier transform operator by $\mathcal F$ and frequency by $f$.
%
Define $R_{n,m}(f)$ as the Fourier transform of $r_{n,m}$ at the frequency $f$,
%
\begin{equation}
\label{dft1}
\begin{split}
 \ R_{n,m}(f):= & \mathcal F\{r_{n,m}(t)\} (f) \\
=& \sum_{k=1}^K\sum_{l=1}^K Q_{lk}(f)\cdot e^{2\pi \si f (\tau_{n,l} - \tau_{m,k})}.
\end{split}
\end{equation}
where
\begin{eqnarray*}
	Q_{lk}(f) = \mathcal F \{
s_l(t) \otimes s_k(t)
\}(f) =  S_l^*(f)  S_k(f).
\end{eqnarray*}
The matrix $Q(f)$ depends on unknown signal waveforms $s_k(t)$ as well as the the number of sources $K$. For Gaussian signals defined in (\ref{model}), we can write specifically
\begin{equation*}
\begin{split}
	& R_{n,m}(f) = \sum_{k=1}^K\sum_{l=1}^K Q_{lk}(f)\cdot e^{2\pi \si f (\tau_{n,l} - \tau_{m,k})}\\
	& = \sum_{k=1}^K\sum_{l=1}^K \rho_k\rho_l e^{2\pi \si f (\tau_{n,l} - \tau_{m,k})}\pi\sqrt{\frac{1}{\alpha_k\alpha_l}}e^{-\pi^2f^2(\frac{1}{\alpha_k}+\frac{1}{\alpha_l})}.
\end{split}
\end{equation*}

Now we can write ${R}_{n,m}(f)$ in (\ref{dft1}) in a compact form and show its low-rank structure. Define a matrix $Q (f)\in \mathbb C^{K\times K}$, where the $(l, k)$th entry of the matrix is $Q_{lk}(f)$. Clearly, $Q(f)$ is a rank-one complex matrix. Define
\[
S(f) = [ {S_1}^*(f), \ldots, S_K^*(f)]^\top,
\]
then
\[	Q(f) = S(f)S(f)^H,
\]
where $(\cdot)^H$ denote the Hermitian of a complex vector or matrix (i.e., the complex conjugate and transpose).
Define
\begin{eqnarray*}
	\alpha_n = [e^{- 2\pi \si f \tau_{n,1}}, e^{- 2\pi \si f \tau_{n,2}},\ldots,e^{- 2\pi \si f \tau_{n,K}}]^\top.
\end{eqnarray*}
We have
\[R_{n,m}(f) = \alpha^H_n  Q(f)\alpha_m, \forall f.\]
Define a matrix $A = [\alpha_1, \ldots, \alpha_N] \in \mathbb C^{K\times N}$, and a matrix $R(f)$, whose $(n, m)$th entry is given by $R_{nm}(f)$. We can further write
\begin{eqnarray*}
R(f) = A^H Q(f) A, \forall f.
\end{eqnarray*}

Assume our observations are a subset of entries of the tensor $R$ with additive Gaussian noise. The missing data can be due to distance and communication constraints; see \cite{xie2018communication} for context. Certain pairs of cross-correlations functions are not available. This can happen when sensors far away, and it is impractical for them to communicate information and perform cross-correlation, and only a subset of frequency samples are communicated. This can also happen when the signal-to-noise ratio is too small for a pair of sensors. Denote the indices of the observations as $\Omega$. To recap, our goal is to infer $K$, from noisy and partial observations of a complex tensor $R$, indexed on $\Omega$.

Now we present the form of the non-linear map. Consider discrete-time and frequency samples. Assume the discrete event samples are indexed by $t = 0, \ldots, T-1$. Thus, for discrete Fourier transform, the frequency samples are also indexed by $f = 0, \ldots, T-1$.
Define a vector of coefficients in our problem $\xi \in \Xi\subset\mathbb R^{2K+NK}$: \[\xi = (\rho_1,\dots,\rho_K, \alpha_1,\dots, \alpha_K, \tau_{1,1}, \tau_{1, 2}, \ldots, \tau_{N, K}).\]
Define the set \[\mathcal L=\{M\in\mathbb R^{N\times N\times T}: M_{i, j, k}=0,\forall (i, j, k)\in \Omega\},\] which can be viewed as the ``nullspace'' of a given observation index set $\Omega$. Then we set \[\theta = (\xi, M_1, M_2),\]  where $M_1 \in \mathcal L$ and $M_1 \in \mathcal L$.
Denote the real and imaginary parts of the frequency samples as $\mathcal R_{n,m,f} = {\rm Re}( R_{n,m}(f))$, and
	$\mathcal I_{n,m,f} = {\rm Im}(R_{n,m}(f))$, respectively, and define the corresponding tensors $\mathcal R$ and $\mathcal I$ (which depend on the parameter vector $\xi$). The non-linear map (similar to the case the complex matrix completion) is defined by
	\begin{equation}
	G(\theta) :=
	(\mathcal R + M_1, \mathcal I + M_2).
	\label{mapG}
	\end{equation}
Hence, although the situation is fairly complex here, we can cast it into the format of the general problem and use our result.

Numerical experiments suggest the following formula for the
characteristic rank $$\ccr=2K+NK-1.$$ This is achieved by evaluating the rank of the Jacobian matrix of the map defined by (\ref{mapG}) (see Remark \ref{rem-rank}) and the appendix for the derivation of the Jacobian matrix).

%
	
	



\section{Numerical Experiments}\label{sec:eg}
\subsection{Complex matrix completion}

In this section  we consider the complex matrix completion problem \eqref{gen3}. To solve the related optimization problem, we use a generalize version the hard thresholding algorithm in \cite{mazumder2010spectral}. In the experiment, we generate a rank-$r$ complex matrix with size $n_1\times n_2$, by first generating $V_1, V_2\in\bbr^{n_1\times r}$ and $W_1, W_2\in\bbr^{n_2\times r}$, where each entries are i.i.d $\N(0,1)$,  and form  $X = (V_1+\si V_2)(W_1+\si W_2 )^\top.$ We numerically verified that the characteristic rank of the manifold $\M_r\subset \bbc^{n_1\times n_2}$, of matrices of rank $r$,  is $\rho=2r(n_1+n_2-r)$ for all random instances, which is consistent with the results in Section \ref{complex}.



\yao{To show the asymptotic distribution of test statistics (Theorem \ref{th-asym}), we generate a rank-2 true matrix $X^*\in\bbc^{100\times 100}$. The observed entries are contaminated with Gaussian noise:
\[
Y_{ij} = X^*_{ij} + \varepsilon_{ij}^{(k)} +\si\eta_{ij}^{(k)}, \; (i,j)\in\Omega,
\]
where $|\Omega| = 1500$ and the noise $\varepsilon_{ij}^{(k)}, \eta_{ij}^{(k)}\overset{iid}{\sim} \mathcal N(0,5^2)$. The experiments are repeated 400 times, i.e., $k = 1,\dots, 400$, to demonstrate the empirical distribution of the test statistic. Figure \ref{com_qq} shows the QQ-plot of $\{T_N(2)^{(k)}\}_{k=1}^{400}$ against the $\chi^2$ distribution with a degrees-of-freedom equal to 2208. Recall that the characteristic rank of the manifold $\M_r\subset \bbc^{n_1\times n_2}$, of matrices of rank $r$, is $\rho=2r(n_1+n_2-r)$ (see Section \ref{complex}). The results in Figure \ref{com_qq} show that the $\chi^2$ distribution fits the test statistics reasonably well.  Moreover, we show the result of detecting the rank in table \ref{exp-compRank}, with the same experiment setting. In each experiment, we complete the matrix from rank $r = 1$ to $r = 4$. We choose the smallest $r$, such that $T_N(r)$ has $p$-value larger than 0.05. In table \ref{exp-compRank}, there are the results of 200 experiments for true rank $r^* = 2$ and $r^* = 3$. We can see the power of tests are high when $r<r^*$ since there is no false acceptance and the false rejection rate is close to the significant level 0.05 when $r=r^*$.}
\begin{table*}[h]
	\centering
	\caption{\yao{Result of hypothesis tests for the rank of complex matrix completion: $r^*$ is the true rank. For each $r^*$, there are 200 experiments. We perform the test from $r=1$ to $r = 4$ and count the number of determined $r$ with significant level, 0.05; $r=0$ means tests are rejected for $r = 1,\dots, 4$.}}
	\vspace{.1in}
	\begin{tabular}{c|ccccc|c}
	\hline
		& $r=0$&$r=1$&$r=2$& $r=3$& $r=4$&FDR\\ \hline
		$r^*=2$&0& 0&190&10&0&$5\%$\\
		$r^*=3$&0&0&0&193&7&$3.5\%$\\\hline
	\end{tabular}
	\label{exp-compRank}
\end{table*}

 \begin{figure}[!h]
 	\centering
 	\includegraphics[width = 0.4\textwidth]{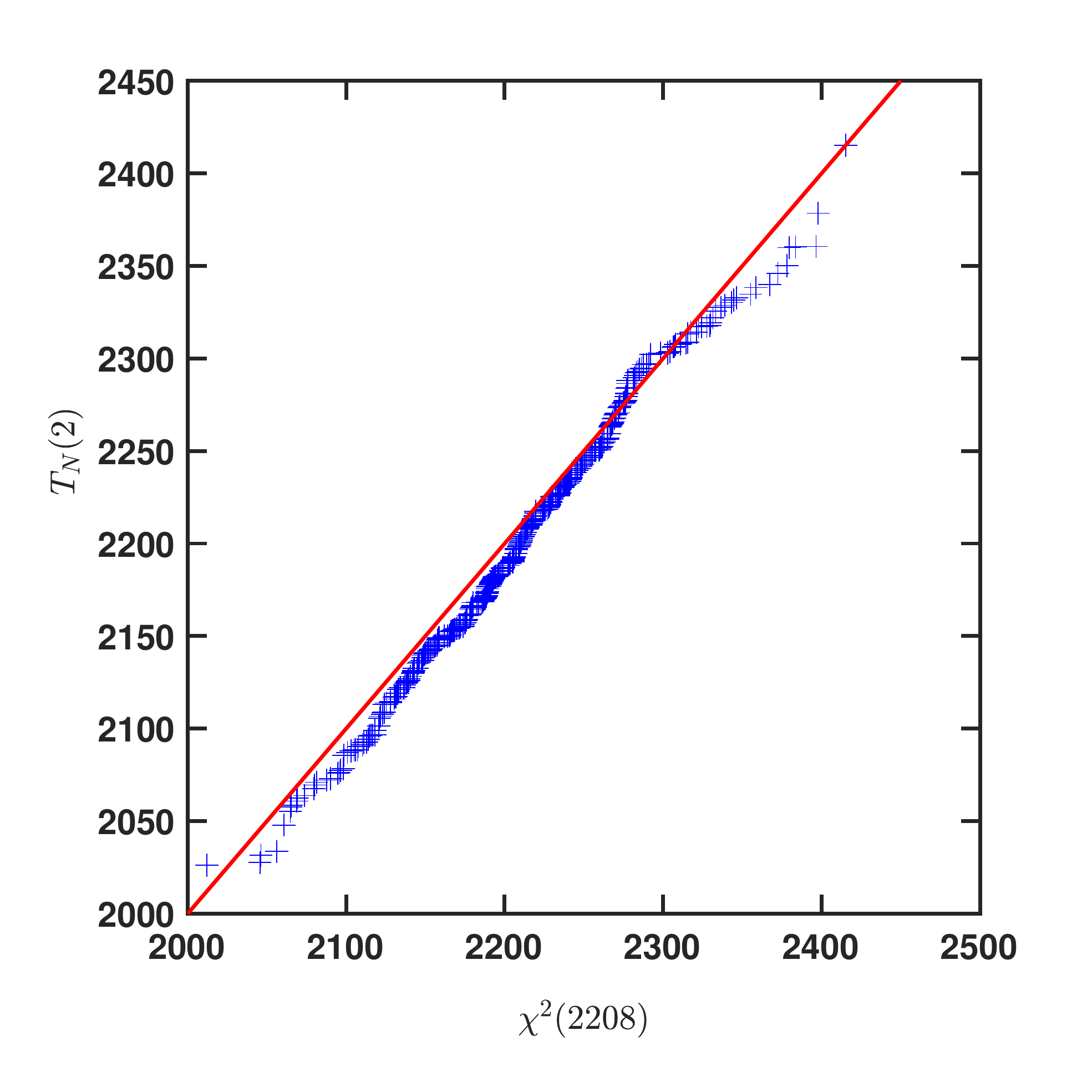}
 \caption{QQ-plot of test statistics against $\chi^2$ distribution.}
 \label{com_qq}
 \end{figure}

\subsection{Characteristic rank of third order tensor}
To generate  third-order tensors of size $n_1\times n_2\times n_3$, we  form $A\in\bbr^{n_1\times r}$, $B\in\bbr^{n_2\times r}$, $C\in\bbr^{n_3\times r}$, where each entry in $A$, $B$, $C$ are {\it i.i.d.} distributed as standard normal (zero-mean and unit variance). Let $X = A\otimes B\otimes C$ and $a^k$, $b^k$, $c^k$ be  the $k$th columns of $A$, $B$, $C$, respectively. To compute the Jacobian matrix, for all $i=1,\dots,n_1$, $j=1,\dots, n_2$, $l = 1,\dots, n_3$ and $k=1,\dots, r$, we can show that
\begin{eqnarray*}
	\frac{\partial X_{ijl}}{\partial a^k_i} = b^k_jc^k_l,\,~ \frac{\partial X_{ijl}}{\partial b^k_j} = a^k_ic^k_l, \, ~\frac{\partial X_{ijl}}{\partial c^k_l} = a^k_ib^k_j.
\end{eqnarray*} All the other entries in the Jacobian matrix are zero.

Table \ref{tensor-table} shows the rank (evaluated numerically) of the Jacobian matrices for different $(n_1,n_2,n_3, r)$ values. We note that when $r$ is sufficiently small, the characteristic rank is equal to $r(n_1+n_2+n_3-2)$, as expected. When $r$ is large, the characteristic rank can be less than $r(n_1+n_2+n_3-2)$. This effect can be explained by Proposition \ref{pr-char}: since in those cases the model is not generically locally identifiable, and hence is not generically identifiable.
It is not surprising that when $r$ is large enough (the cases marked  with * in the left column),  the rank of the Jacobian matrix is equal to $n_1n_2n_3$. The interesting cases are when $r \approx (n_1n_2n_3)/(n_1+n_2+n_3-2)$. The right column of table \ref{tensor-table} shows some cases in  which  ranks of the Jacobian matrices are less than $\min\{n_1n_2n_3, r(n_1+n_2+n_3-2)\}$.

\begin{table}[h]
\centering
	\caption{Rank of the Jacobian matrices for third order tensor. For each combination of $(n_1,n_2,n_3,r)$, the experiments are repeated 100 times and the results are all the same. When $r$ is small, rank$(J) = r(n_1+ n_2+n_3-2)$. When $r$ is large (cases marked  with $^*$), rank$(J) < r(n_1+n_2+n_3-2)$.}
	\vspace{.1in}
	{\small
	\begin{tabular}{ccccc|ccccc}
	\hline
		$n_1$& $n_2$&$n_3$&$r$&rank($J$)& $n_1$& $n_2$&$n_3$&$r$&rank($J$)\\ \hline
		3&4&5&1&10&2&2&4&3&15$^*$\\
		3&4&5&5&50&2&2&5&3&18$^*$\\
		3&4&5&12&60$^*$&2&3&5&4&28$^*$\\
	15&15&15&5&215&3&3&3&4&26$^*$\\
		15&15&15&15&645&3&4&4&5&44\\
		15&15&15&100&3375$^*$&3&5&5&7&74$^*$\\\hline
	\end{tabular}
	}
	\label{tensor-table}
\end{table}

\subsection{Determining the number of signals in blind de-mixing}

Consider the ambient noise imaging in a distributed sensor network setting (described in Section \ref{seismic}), where there are missing values in the observations.  Our goal is to determine the number of sources. For this problem, one can show that the characteristic rank is $2K+NK-1$ for large enough $T$. Therefore, by identifying the characteristic rank, we can determine the number of sources.

In each experiment, we generate the random instances are follows: $\alpha_k\sim \mbox{Unif}[10,11]$, $\rho_k\sim \mbox{Unif}[10,11]$, $\tau_{n,k}\sim \mbox{Unif}[-2.5, 2.5]$, $\forall n = 1,\dots N$ and $k = 1,\dots,K^*$.

First, we want to verify the characteristic rank of the Jacobian matrix predicted using our theory. Let $N= 8, 10 ,12$ and $K = 1,\dots,5$. For each $N$ and $K$, we generate parameters and compute the corresponding rank of the Jacobian matrix numerically. In figure \ref{seismic-crank}, each point is the mean of ranks in 100 experiments corresponding to a certain pair of $N$ and $K$. The lines plotted correspond to $2K+NK-1$, for $N = 8, 10,12$. We can see the points are exactly on the lines, which justifies our formulation for the characteristic rank.

Second, we show the result of testing the rank in this problem. The observation noise are normal random variables with zero mean and variance equal to 0.05. Table \ref{rtest1} is the result of determining source number $K^*$ with $\alpha_k$, $\rho_k$ and $\tau_{n,k}$ being unknown. We run experiments for $K^*=1,\dots,5$. For each $K^*$, 100 experiments are run and in each experiment, the test is running from $K = 1$ to $K = 6$ and the significant level is $0.01$. In the table, $K=0$ means all the tests are rejected. We can see our test gives the true number of sources most of the time, except $K^* = 5$. When $K^* = 5$, the algorithm becomes difficult to converge to the optimal solution and therefore leads to a large fitting error.

\begin{figure}[!h]
\centering
\includegraphics[width = 0.3\textwidth]{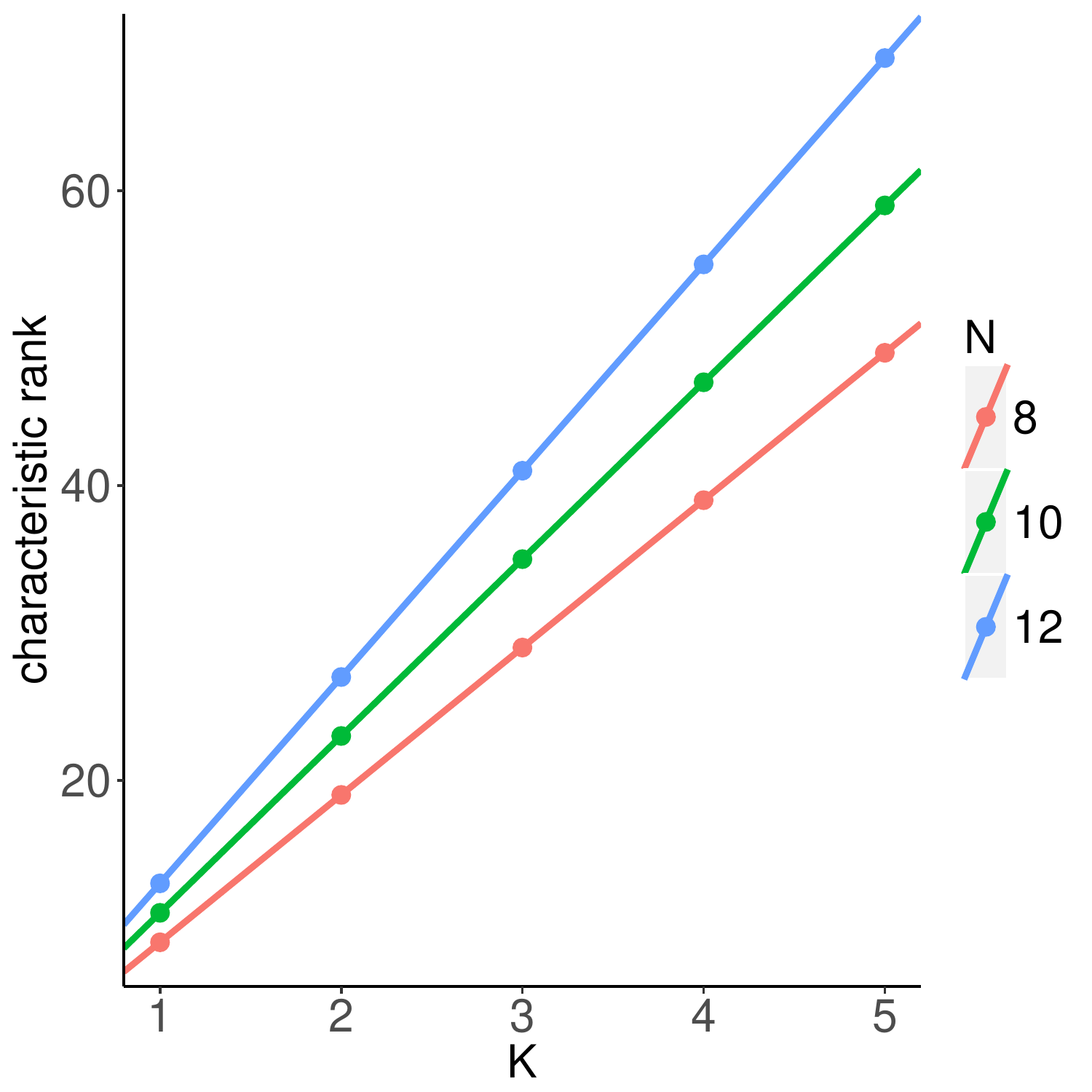}
\caption{The characteristic rank of the problem in Section \ref{seismic}: $K$ is the number of sources, $N$ is the number of sensors, the points are the rank of the Jacobian matrix of the mapping, and the line is $2K + NK -1$.}
	\label{seismic-crank}
 \end{figure}

\begin{table*}[h]
\centering
	\caption{Results of hypothesis tests for the number of sources: $K^*$ is the true number of sources. For each $K^*$, there are 100 experiments. We perform the test from $K=1$ to $K=6$ and count the number of determined $K$; $K=0$ means tests are rejected for $K=1,\dots,6$.}
	\vspace{.1in}
	\begin{tabular}{c|ccccccc|c}
	\hline
		& $K=0$&$K=1$&$K=2$& $K=3$& $K=4$& $K=5$& $K=6$&FDR\\ \hline
		$K^*=1$&0& 100&0&0&0&0&0&0\\
		$K^*=2$&1&0&98&0&1&0&0&$6\%$\\
		$K^*=3$&4&0&0&94&2&0&0&$3\%$\\
		$K^*=4$&9&0&0&0&91&0&0&$9\%$\\
		$K^*=5$&32&0&0&0&0&67&1&$33\%$\\ \hline
	\end{tabular}
	\label{rtest1}
\end{table*}

\subsection{One-hidden-layer neural networks}

In this section, we consider the problem of determining the number of hidden units for one-hidden-layer neural networks; the problem described in (\ref{ohl-ls}). In the experiment, $x_i\sim\N(0, I_{d})$, $U\in\bbr^{d\times r^*}$, such that $U_{ij}\sim\N(0,1)$ and $m=1000$. Consider the activation function to be quadratic activation and sigmoid activation, respectively. Table \ref{quad-table} and Table \ref{sigmoid-table} are the ranks of Jacobian matrices for different combinations of $(d,r^*)$. The results justify the formula of characteristic rank of one-hidden-layer neural networks are $dr^* - r^*(r^*-1)/2$ for quadratic activation and $dr^*$ for sigmoid activation, respectively.

Although we could not provide any theoretical prediction for the characteristic rank when the activation function is a ReLu function, here we provide some numerical examples. We show the performance of our rank test for one-hidden-layer neural networks with a ReLU activation function. In the experiments, $d = 50$, and $\sigma = 0.1$. We perform 100 experiments each from $r^*$, with the true rank of $U$ being equals to 1 to 6. For each $r^*$, we perform the test from $r=1$ to $r=7$ with significant level 0.05. With this setting, the p-value is computed under the $\chi^2(m - dr)$. The optimization problem involved with fitting the neural networks model is solved using gradient descent (implemented by \textsf{Pytorch} package).

For ReLu activation function, Table \ref{rtest2} shows the rank determined by our proposed test for each $r^*$. Here, $r=0$ means all tests are rejected. Results are similar to what we observed in Table \ref{rtest1}. When the order of the model is small, the test is consistent with the significant level. When the order of the model increase, convergence to the optimal solution becomes more difficult; in this setting, the false discovery rate will increase but is still tolerable. An interesting finding is that our test still gives promising results even though the ReLU activation is not an analytic function. \begin{table}[h!]
\centering
		\caption{Rank of the Jacobian matrix for one-hidden-layer neural networks with a quadratic activation function. For each combination of $(d,r^*)$, the experiments are repeated 100 times, and the results are all the same. This justifies the formula of the characteristic rank of one-hidden-layer neural networks with quadratic activation is $dr^* - r^*(r^*-1)/2$.}
		\vspace{.1in}
		\begin{tabular}{ccc|ccc}
	\hline
		$d$& $r^*$&rank($J$)& $d$&$r^*	$&rank($J$)\\ \hline
		10&1&10&30&11&275\\
		10&5&40&30&17&374\\
		10&10&55&30&23&473\\
		20&1&20&50&10&455\\
		20&12&174&70&10&655\\
		20&18&207&90&10&855\\ \hline
	\end{tabular}
	\label{quad-table}
\end{table}

 \begin{table}[h]
\centering
	\caption{Rank of the Jacobian matrix for one-hidden-layer neural networks with sigmoid activation. For each combination of $(d,r^*)$, the experiments are repeated 100 times and the results are all the same. This justifies the formula of the characteristic rank of one-hidden-layer neural networks with sigmoid activation is $dr^*$.}
	\vspace{.1in}
	\begin{tabular}{ccc|ccc}
	\hline
		$d$& $r^*$&rank($J$)& $d$&$r^*	$&rank($J$)\\ \hline
		10&1&10&30&11&330\\
		10&5&50&30&17&510\\
		10&10&100&30&23&690\\
		20&1&20&50&10&500\\
		20&12&240&70&10&700\\
		20&18&360&90&10&900\\ \hline
	\end{tabular}
	\label{sigmoid-table}
\end{table}

\begin{table*}[h]
\centering
	\caption{Result of ReLU activation function: $r^*$ is the rank of true $U^*$. For each $r^*$, there are 100 experiments. We perform the test from $r=1$ to $r=7$ and count the number of determined $r$. $r=0$ means tests are rejected for $r=1,\ldots,7$.}
	\vspace{.1in}
	\begin{tabular}{c|cccccccc|c}
	\hline
		& $r=0$&$r=1$&$r=2$& $r=3$& $r=4$& $r=5$& $r=6$& $r=7$&FDR\\ \hline
		$r^*=2$&3& 0&96&1&0&0&0&0&$4\%$\\
		$r^*=3$&4&0&0&96&0&0&0&0&$4\%$\\
		$r^*=4$&4&0&0&0&94&1&0&1&$6\%$\\
		$r^*=5$&2&0&0&0&0&93&5&0&$7\%$\\
		$r^*=6$&5&0&0&0&0&0&88&7&$12\%$\\ \hline
	\end{tabular}
	\label{rtest2}
\end{table*}

\section{Conclusions}\label{sec:conclusion}

We develop a general theory for the goodness-of-fit test to non-linear models, which essentially shows that the parameter-of-interests are related to the characteristic rank of the linear map that defines the manifold structure of our observation. The test statistic has a simple chi-square distribution whose parameters are specified explicitly. Based on this result, it is convenient to implement a test procedure to determine the model order in practice. Our general theory can provide precise answers to several questions, such as determining the rank of (complex) low-rank matrix from noisy and incomplete observations. In some other applications, we show that how the general theory can shed light on finding the ``model-order-of-interests'', such as tensor completion, determining the number of hidden nodes in neural networks, determining the number of sources in blind signal demixing problems, using analysis and simulations. Providing explicit answers (such as exact values of characteristic ranks) are too complex and beyond the scope of this paper, which we leave for future work.

\bibliography{References}

\begin{thebibliography}{10}
\providecommand{\url}[1]{#1}
\csname url@samestyle\endcsname
\providecommand{\newblock}{\relax}
\providecommand{\bibinfo}[2]{#2}
\providecommand{\BIBentrySTDinterwordspacing}{\spaceskip=0pt\relax}
\providecommand{\BIBentryALTinterwordstretchfactor}{4}
\providecommand{\BIBentryALTinterwordspacing}{\spaceskip=\fontdimen2\font plus
\BIBentryALTinterwordstretchfactor\fontdimen3\font minus
  \fontdimen4\font\relax}
\providecommand{\BIBforeignlanguage}[2]{{%
\expandafter\ifx\csname l@#1\endcsname\relax
\typeout{** WARNING: IEEEtran.bst: No hyphenation pattern has been}%
\typeout{** loaded for the language `#1'. Using the pattern for}%
\typeout{** the default language instead.}%
\else
\language=\csname l@#1\endcsname
\fi
#2}}
\providecommand{\BIBdecl}{\relax}
\BIBdecl

\bibitem{ding2018model}
J.~Ding, V.~Tarokh, and Y.~Yang, ``Model selection techniques: An overview,''
  \emph{IEEE Signal Processing Magazine}, vol.~35, no.~6, pp. 16--34, 2018.

\bibitem{Test05}
E.~Lehmann and J.~Romano, \emph{Testing Statistical Hypotheses}.\hskip 1em plus
  0.5em minus 0.4em\relax Springer, 2005.

\bibitem{BrockwellDavis2010}
P.~Brockwell and R.~Davis, \emph{Introduction to Time Series and
  Forecasting}.\hskip 1em plus 0.5em minus 0.4em\relax Springer, 2010.

\bibitem{hastie2005elements}
T.~Hastie, R.~Tibshirani, J.~Friedman, and J.~Franklin, ``The elements of
  statistical learning: data mining, inference and prediction,'' \emph{The
  Mathematical Intelligencer}, vol.~27, no.~2, pp. 83--85, 2005.

\bibitem{chwialkowski2016kernel}
K.~Chwialkowski, H.~Strathmann, and A.~Gretton, ``A kernel test of goodness of
  fit.''\hskip 1em plus 0.5em minus 0.4em\relax JMLR: Workshop and Conference
  Proceedings, 2016.

\bibitem{liu2016kernelized}
Q.~Liu, J.~Lee, and M.~Jordan, ``A kernelized stein discrepancy for
  goodness-of-fit tests,'' in \emph{International conference on machine
  learning}, 2016, pp. 276--284.

\bibitem{jitkrittum2017linear}
W.~Jitkrittum, W.~Xu, Z.~Szab{\'o}, K.~Fukumizu, and A.~Gretton, ``A
  linear-time kernel goodness-of-fit test,'' in \emph{Advances in Neural
  Information Processing Systems}, 2017, pp. 262--271.

\bibitem{verdinelli1998bayesian}
I.~Verdinelli, L.~Wasserman \emph{et~al.}, ``Bayesian goodness-of-fit testing
  using infinite-dimensional exponential families,'' \emph{The Annals of
  Statistics}, vol.~26, no.~4, pp. 1215--1241, 1998.

\bibitem{fefferman2016testing}
C.~Fefferman, S.~Mitter, and H.~Narayanan, ``Testing the manifold hypothesis,''
  \emph{Journal of the American Mathematical Society}, vol.~29, no.~4, pp.
  983--1049, 2016.

\bibitem{candes2010power}
E.~J. Cand{\`e}s and T.~Tao, ``The power of convex relaxation: Near-optimal
  matrix completion,'' \emph{IEEE Trans. Info. Theory}, vol.~56, no.~5, pp.
  2053--2080, 2010.

\bibitem{candes2010matrix}
E.~J. Candes and Y.~Plan, ``Matrix completion with noise,'' \emph{Proceedings
  of the IEEE}, vol.~98, no.~6, pp. 925--936, 2010.

\bibitem{recht2010guaranteed}
B.~Recht, M.~Fazel, and P.~A. Parrilo, ``Guaranteed minimum-rank solutions of
  linear matrix equations via nuclear norm minimization,'' \emph{SIAM Review},
  vol.~52, no.~3, pp. 471--501, 2010.

\bibitem{klopp2014noisy}
O.~Klopp \emph{et~al.}, ``Noisy low-rank matrix completion with general
  sampling distribution,'' \emph{Bernoulli}, vol.~20, no.~1, pp. 282--303,
  2014.

\bibitem{DavenportRomberg16}
M.~Davenport and J.~Romberg, ``An overview of low-rank matrix recovery from
  incomplete observations,'' \emph{IEEE Journal of Selected Topics in Signal
  Processing}, vol.~10, no.~4, pp. 608--622, 2016.

\bibitem{pimentel2016characterization}
D.~L. Pimentel-Alarc{\'o}n, N.~Boston, and R.~D. Nowak, ``A characterization of
  deterministic sampling patterns for low-rank matrix completion,'' \emph{IEEE
  Journal of Selected Topics in Signal Processing}, vol.~10, no.~4, pp.
  623--636, 2016.

\bibitem{sxz19}
A.~Shapiro, Y.~Xie, and R.~Zhang, ``Matrix completion with deterministic
  pattern - a geometric perspecve,'' \emph{IEEE Transactions on Signal
  Processing}, vol.~67, pp. 1088--1103, 2019.

\bibitem{hastie2015matrix}
T.~Hastie, R.~Mazumder, J.~D. Lee, and R.~Zadeh, ``Matrix completion and
  low-rank svd via fast alternating least squares,'' \emph{The Journal of
  Machine Learning Research}, vol.~16, no.~1, pp. 3367--3402, 2015.

\bibitem{recht2013parallel}
B.~Recht and C.~R{\'e}, ``Parallel stochastic gradient algorithms for
  large-scale matrix completion,'' \emph{Mathematical Programming Computation},
  vol.~5, no.~2, pp. 201--226, 2013.

\bibitem{sha1986}
A.~Shapiro, ``Asymptotic theory of overparameterized structural models,''
  \emph{Journal of the American Statistical Association}, vol.~81, pp.
  142--149, 1986.

\bibitem{stern}
S.~Sternberg, \emph{Lectures on differential geometry}.\hskip 1em plus 0.5em
  minus 0.4em\relax Englewood Cliff: Prentice Hall, Inc., 1964.

\bibitem{sards1942measure}
A.~Sards, ``The measure of critical values of differential maps,'' \emph{Bull.
  Amer. Math. Soc}, vol.~48, pp. 883--890, 1942.

\bibitem{fed69}
H.~Federer, \emph{Geometric Measure Theory}.\hskip 1em plus 0.5em minus
  0.4em\relax New York: Springer Verlag, 1969.

\bibitem{McManus}
D.~McManus, ``Who invented local power analysis?'' \emph{Econometric Theory},
  vol.~7, pp. 265--268, 1991.

\bibitem{SDR2014}
A.~Shapiro, D.~Dentcheva, and A.~Ruszczy{\'n}ski, \emph{Lectures on
  {S}tochastic {P}rogramming}, 2nd~ed., ser. MOS-SIAM Series on
  Optimization.\hskip 1em plus 0.5em minus 0.4em\relax SIAM, 2014.

\bibitem{ste1985}
J.~Steiger, A.~Shapiro, and M.~Browne, ``On the multivariate asymptotic
  distribution of sequential chi-square statistics,'' \emph{Psychometrika},
  vol.~50, pp. 253--254, 1985.

\bibitem{li2017algorithmic}
Y.~Li, T.~Ma, and H.~Zhang, ``Algorithmic regularization in over-parameterized
  matrix recovery,'' \emph{arXiv preprint arXiv:1712.09203}, 2017.

\bibitem{kolda2009tensor}
T.~G. Kolda and B.~W. Bader, ``Tensor decompositions and applications,''
  \emph{SIAM review}, vol.~51, no.~3, pp. 455--500, 2009.

\bibitem{chian2017}
L.~Chiantini, G.~Ottaviani, and N.~Vannieuwenhoveni, ``Effective criteria for
  specific identifiability of tensors and forms,'' \emph{SIAM Journal on Matrix
  Analysis and Applications}, vol.~38, pp. 656--681, 2017.

\bibitem{doman}
I.~Domanov and L.~D. Lathauwer, ``Generic uniqueness conditions for the
  canonical polyadic decomposition and {INDSCAL},'' \emph{SIAM Journal on
  Matrix Analysis and Applications}, vol.~36, pp. 1567--1589, 2015.

\bibitem{7919265}
S.~{Ling} and T.~{Strohmer}, ``Blind deconvolution meets blind demixing:
  Algorithms and performance bounds,'' \emph{IEEE Transactions on Information
  Theory}, vol.~63, no.~7, pp. 4497--4520, July 2017.

\bibitem{snieder2010imaging}
R.~Snieder and K.~Wapenaar, ``Imaging with ambient noise,'' \emph{Physics
  Today}, vol.~63, no.~9, pp. 44--49, 2010.

\bibitem{mccoy2014sharp}
M.~B. McCoy and J.~A. Tropp, ``Sharp recovery bounds for convex demixing, with
  applications,'' \emph{Foundations of Computational Mathematics}, vol.~14,
  no.~3, pp. 503--567, 2014.

\bibitem{xie2018communication}
L.~Xie, Y.~Xie, S.-M. Wu, F.-C. Lin, and W.~Song, ``Communication efficient
  signal detection for distributed ambient noise imaging,'' in \emph{2018 52nd
  Asilomar Conference on Signals, Systems, and Computers}.\hskip 1em plus 0.5em
  minus 0.4em\relax IEEE, 2018, pp. 1779--1783.

\bibitem{mazumder2010spectral}
R.~Mazumder, T.~Hastie, and R.~Tibshirani, ``Spectral regularization algorithms
  for learning large incomplete matrices,'' \emph{Journal of machine learning
  research}, vol.~11, no. Aug, pp. 2287--2322, 2010.

\bibitem{fish1966}
F.~M. Fisher, \emph{The identification problem in econometrics}.\hskip 1em plus
  0.5em minus 0.4em\relax New York: McGraw-Hill Company, 1966.

\end{thebibliography}

\section{Appendix}
\label{sec-appen}


\noindent{\bf Proof of Proposition \ref{pr-1}}
(i) Since $G(\cdot)$ is twice continuously differentiable, it follows that
$J(\cdot)$ is continuous. Thus    the function
$ \rank (J(\cdot))$ is lower semicontinuous, and hence  the set
$\left \{\theta\in \Theta: \rank (J(\theta))\le \ccr-1\right\}$  is closed. It follows that  its complement    set $\left \{\theta\in \Theta: \rank (J(\theta))=\ccr\right\}$  is open.

(ii)
Let $\theta_0\in \Theta$ be such that $\rank(J(\theta_0))=\ccr$, such $\theta_0$ exists since the function $ \rank (J(\cdot))$ is piecewise constant.  Consider an $\ccr\times \ccr$ submatrix of $J(\theta_0)$ of rank $\ccr$,  and the associated  function $\phi(\theta)$ given by the determinant of this submatrix of $J(\theta)$. Since $G(\cdot)$ is analytic, we have that the function $\phi(\cdot)$ is analytic  and is not constantly zero since  $\phi(\theta_0)\ne 0$.
It follows that the set $\{\theta:\phi(\theta)=0\}$ has (Lebesgue)  measure zero (e.g., \cite{fish1966}). That is, for a.e. $\theta$ we have that $\rank (J(\theta))\ge \ccr$. Since by the definition the rank $\ccr$ is maximal, it  follows that  $\rank(J(\theta))=\ccr$ for  a.e.   $\theta\in \Theta$.
  This completes the proof.
\\
\\
{\bf Proof of Proposition \ref{pr-uniq}}
Since $\cM$ is a smooth manifold near $x_0$ it can be defined  by equations $\phi(x)=0$ in a neighborhood    of  $x_0$  with $\phi:\bbr^m\to\bbr^m$ being a smooth  near $x_0$ mapping with nonsingular Jacobian matrix $\nabla \phi(x_0)$. Then optimality condition \eqref{nescon} can be written as: there exists $\lambda\in \bbr^m$ such that the derivatives of the Lagrangian $L(x,\lambda):=\half\|\hat{y}-x\|^2- \lambda^\top \phi(x)$ are zeros at $(\hat{x},\lambda)$. This can be written as the following system of equations in $(x,\lambda)$,
\begin{equation}\label{syseq}
\nabla_x L(x,\lambda)=0,\;
 \phi(x)=0.
\end{equation}
Note that as $\hat{y}$ and $x$ approach $x_0$, the corresponding $\lambda$ tends to 0. The
 Jacobian matrix of partial derivatives of  this system, with respect to $(x,\lambda)$, at $x=x_0$ and $\lambda=0$  is
$\begin{psmallmatrix}
 I_m& \nabla  \phi(x_0)  \\
  \nabla \phi(x_0) ^\top  & 0
\end{psmallmatrix}$.
This Jacobian matrix is nonsingular.  It follows by the Implicit Function Theorem that in a neighborhood $\W$  of $x_0$ the system \eqref{syseq} has unique solution. Moreover by Remark \ref{rem-con}  the neighborhood $\W$ can be such that if $\hat{y}\in \W$, then any optimal solution of the least squares problem is in $\W$. If moreover $\hat{x}$ is in $\W$ and satisfies optimality equations \eqref{syseq}, then by the uniqueness property $\hat{x}$  should coincide with the corresponding optimal solution. This completes the proof.
 \\
 \\
 {\bf Proof of Theorem \ref{th-asym}}
Since  $\hat{y}$ converges in probability to $x_0$,
the assertion (i) follows from Proposition \ref{pr-uniq}.
Also any minimizer $\hat{x}$ in the right hand side of \eqref{mod-2} converges in probability to $x_0$ (see Remark \ref{rem-con}).
 Therefore we can perform the asymptotic analysis in  a neighborhood of $x_0$.
As in the above proof of Proposition \ref{pr-uniq},  $\cM$  can be defined  by   equations $\phi(x)=0$ in a neighborhood    of  $x_0$  with  nonsingular Jacobian matrix $\nabla \phi(x_0)$. Let $(\hat{x},\hat{\lambda})$ be a solution of equations \eqref{syseq} in a sufficiently small  neighborhood of $(x_0,0)$. By the  Implicit Function Theorem we have that
\begin{equation}\label{impth}
\begin{split}
  \left[\begin{array}{ccc}
   \hat{x}-x_0\\
   \hat{\lambda}
    \end{array}
   \right] &=
  \left[\begin{array}{ccc}
  I_m& \nabla  \phi(x_0)  \\
 \nabla \phi(x_0) ^\top  & 0
 \end{array}\right] ^{-1}
  \left[\begin{array}{ccc}
   \hat{y}-x_0\\
   0
    \end{array}
   \right]\\
   &\quad+o(\| \hat{y}-x_0\|).
   \end{split}
\end{equation}
Also it follows by \eqref{mod-1} that $N^{1/2}(\hat{y}-x_0)$ converges in distribution to normal $\N(\gamma,\sigma^2 I_m)$.  In particular this implies that $\| \hat{y}-x_0\|=O_p(N^{-1/2})$, and hence
\begin{equation}\label{imp-2}
 \hat{x}-x_0=P
 ( \hat{y}-x_0)+o_p(N^{-1/2}),
\end{equation}
where
\begin{equation}\label{imp-3}
P= I_m- \nabla  \phi(x_0)\left ( \nabla  \phi(x_0)^\top \nabla  \phi(x_0) \right )^{-1}
 \nabla  \phi(x_0)^\top.
\end{equation}
Note that   $\T_\cM(x_0)=\{v:\nabla  \phi(x_0)^\top v=0\}$.  Therefore matrix $P$  in \eqref{imp-3}  is the orthogonal projection matrix onto the  tangent space $\T_\cM(x_0)$.   Slutsky's theorem together with \eqref{imp-2} imply that  $N^{1/2}(\hat{x}-x_0)$ has the same asymptotic distribution as
$ P[N^{1/2} ( \hat{y}-x_0)]$. Since $N^{1/2}(\hat{y}-x_0)$ converges in distribution to normal $\N(\gamma,\sigma^2 I_m)$, the assertion (iii) follows, and the  assertion (iv) follows by similar arguments.

Moreover by  \eqref{imp-2},
\[
  \hat{y}- \hat{x} = \hat{y}-x_0- ( \hat{x} -x_0)=
 (I_m- P) ( \hat{y}-x_0)+o_p(N^{-1/2}),
 \]
and since  $\| \hat{y}-x_0\|=O_p(N^{-1/2})$  it follows that
\begin{equation}\label{imp-4}
 \|\hat{y}- \hat{x}\|_2^2 = \| (I_m- P) ( \hat{y}-x_0)\|_2^2+o_p(N^{-1}).
\end{equation}
 It follows by Slutsky's theorem that the $N$ times   right hand side of \eqref{imp-4} has the same asymptotic distribution as $Z^\top (I_m- P) Z$, where $Z\sim \N(\gamma,\sigma^2 I_m)$. The assertion (ii) follows. 
This completes the proof.
 \\
 \\
   Theorem  \ref{th-nest} can be proved in a similar way by showing that asymptotically this is equivalent to the linear case.
  \\
  \\
 {\bf Proof of Proposition \ref{pr-wpos}}
Let $x=\G(\xi)$ be a well-posed point.  Then $\T_\M(x)=\{d\G(\xi)h:h\in \bbr^d\}$, and
 for any $\zeta\in \bbr^k$  we have by \eqref{diff} that  dimension of  the image of the differential $dG(\xi,\zeta)$ is $\rho +k$. It follows that  $ \ccr\ge  \rho+k$. Since $ \ccr\le  \rho+k$, it follows that $\ccr=  \rho+k$.

 Conversely  suppose that $\M$ is a smooth manifold of dimension $\rho$ and
  $\ccr= \rho+k$. Let $\theta\in \Theta$ be such that dimension of the image of $dG(\theta)$ is $\ccr$, by Proposition \ref{pr-1} we have that a.e. $\theta$ is like that. Since  $\ccr= \rho+k$ and $\T_\M(x)=\{d\G(\xi)h:h\in \bbr^d\}$ we have  by \eqref{diff} that \eqref{cond} follows.
  It remains to note that $dG(\theta)=dG(\theta')$ for any points $\theta=(\xi,\zeta)$ and $\theta'=(\xi,\zeta')$ in     $\Theta$ with the same first component.  This completes the proof.
\\
\\
\noindent{\bf Proof of Proposition \ref{pr-char}}
Let $\rho$ be the characteristic rank of mapping
\begin{equation}\label{mapping}
 \bbr^{n_1\times r}\times   \bbr^{n_2\times r}\times \bbr^{n_3\times r}  \ni(A,B,C)\mapsto A\otimes B\otimes C.
\end{equation}
Recall that it always holds that $r(n_1+n_2+n_3-2)\ge \rho$.

Consider $\xi=(A,B,C)$ such that rank of the Jacobian matrix of  mapping \eqref{mapping} at $(A,B,C)$ is $\rho$. For $X=A\otimes B\otimes  C$ consider the set
\begin{equation*}
\begin{split}
\G^{-1}(X) & =\left\{(A',B',C')\in \bbr^{n_1\times r}\times   \bbr^{n_2\times r}\times \bbr^{n_3\times r}: \right.\\
& \left. A'\otimes B'\otimes C'=X\right \}.
\end{split}
\end{equation*}
By the Constant Rank Theorem this set forms a smooth manifold
of dimension
\[
\dim\left(\bbr^{n_1\times r}\times   \bbr^{n_2\times r}\times \bbr^{n_3\times r}\right )-\rho=r(n_1+n_2+n_3)-\rho
\]
 in  a neighborhood of the point $\xi$. If \eqref{charrank} holds, then dimension of this manifold is $2r$, and hence any $(A',B',C')\in \G^{-1}(X)$ in a neighborhood of $(A,B,C)$  can be obtained by the rescaling. That is, the local identifiability follows.

On the other hand if $r(n_1+n_2+n_3)-\rho>2r$, then this will imply that there exists  $(A',B',C')\in \bbr^{n_1\times r}\times   \bbr^{n_2\times r}\times \bbr^{n_3\times r}$ near $(A,B,C)$ such that
 $A'\otimes B'\otimes  C'= A\otimes B\otimes  C$ and $(A',B',C')$ cannot be obtained from  $(A,B,C)$ by the rescaling.
 That is, the local identifiability does not hold.

\noindent{\bf Derivation of the Jacobian matrix in section \ref{seismic}.}\\
For all $ k_0 = 1,\dots, K$, $\forall n,m,n_0 = 1,\dots, N$ and $f = 0,\dots, T-1$, the entries of the Jacobian matrix can be derived as follows
\begin{align*}
 \frac{\partial \mathcal R_{n,m,f}}{\partial \rho_{k_0}}
 =& \sum_{l=1}^K \rho_l  (\cos(2\pi f(\tau_{n,l} - \tau_{m,k_0}))\\
& + \cos(2\pi f(\tau_{n,k_0} - \tau_{m,l})))\\
&\cdot \pi\sqrt{\frac{1}{\alpha_{k_0}\alpha_l}}e^{-\pi^2f^2(\frac{1}{\alpha_{k_0}}+\frac{1}{\alpha_l})}.\\ \\
	 \frac{\partial \mathcal I_{n,m,f}}{\partial \rho_{k_0}} 
	 =& \sum_{l=1}^K \rho_l  (\sin(2\pi f(\tau_{n,l} - \tau_{m,k_0}))\\
	&+ \sin(2\pi f(\tau_{n,k_0} - \tau_{m,l})))
	\\
	&\cdot\pi\sqrt{\frac{1}{\alpha_{k_0}\alpha_l}}e^{-\pi^2f^2(\frac{1}{\alpha_{k_0}}+\frac{1}{\alpha_l})}.\\ \\
	\frac{\partial \mathcal R_{n,m,f}}{\partial \alpha_{k_0}} =& -\frac{\pi}{2}\sum_{l=1}^K \rho_{k_0}\rho_l (\cos(2\pi f(\tau_{n,l} - \tau_{m,k_0}))\\
&+\cos(2\pi f (\tau_{n,k_0}- \tau_{m,l})))\\
&\cdot \alpha_{k_0}^{-\frac{3}{2}}\alpha_l^{-\frac{1}{2}}e^{-\pi^2f^2(\frac{1}{\alpha_k}+\frac{1}{\alpha_l})}\\
	&+ \pi^3f^2\sum_{l=1}^K \rho_{k_0}\rho_l  (\cos(2\pi f(\tau_{n,l} - \tau_{m,k_0}))\\
	&+\cos(2\pi f(\tau_{n,k_0} - \tau_{m,l})))\alpha_{k_0}^{-\frac{1}{2}}\\
	&\cdot\alpha_l^{-\frac{1}{2}}e^{-\pi^2f^2(\frac{1}{\alpha_{k_0}}+\frac{1}{\alpha_l})}\alpha_{k_0}^{-2}\\
	=& \frac{\partial \mathcal R_{n,m,f}}{\partial \rho_{k_0}}(-\frac{\rho_{k_0}\alpha_{k_0}^{-1}}{2} +\pi^2f^2\rho_{k_0}\alpha_{k_0}^{-2}).\\ \\
	\frac{\partial \mathcal I_{n,m,f}}{\partial \alpha_{k_0}} =& -\frac{\pi}{2}\sum_{l=1}^K \rho_{k_0}\rho_l (\sin(2\pi f(\tau_{n,l} - \tau_{m,k_0})) \\
	&+\sin(2\pi f (\tau_{n,k_0}- \tau_{m,l})))\\
	&\cdot\alpha_{k_0}^{-\frac{3}{2}}\alpha_l^{-\frac{1}{2}}e^{-\pi^2f^2(\frac{1}{\alpha_k}+\frac{1}{\alpha_l})}\\
	&\pi^3f^2\sum_{l=1}^K \rho_{k_0}\rho_l  (\sin(2\pi f(\tau_{n,l} - \tau_{m,k_0}))\\
	&\sin(2\pi f(\tau_{n,k_0} - \tau_{m,l})))\\
	&\cdot \alpha_{k_0}^{-\frac{1}{2}}\alpha_l^{-\frac{1}{2}}e^{-\pi^2f^2(\frac{1}{\alpha_{k_0}}+\frac{1}{\alpha_l})}\alpha_{k_0}^{-2}\\
	=& \frac{\partial \mathcal I_{n,m,f}}{\partial \rho_{k_0}}(-\frac{\rho_{k_0}\alpha_{k_0}^{-1}}{2} +\pi^2f^2\rho_{k_0}\alpha_{k_0}^{-2}).
\end{align*}
\begin{align*}
	&\frac{\partial \mathcal R_{n,m,f}}{\partial \tau_{n_0,k_0}}\\
	=& \mathbbm{1}(n=n_0)\sum_{l=1}^K \rho_l\rho_{k_0}  (-2\pi f\sin(2\pi f(\tau_{n_0,k_0} - \tau_{m,l})))\\
	& \pi\alpha_l^{-\frac{1}{2}}\alpha_{k_0}^{-\frac{1}{2}}e^{-\pi^2f^2(\frac{1}{\alpha_l}+\frac{1}{\alpha_{k_0}})}\\
	&+\mathbbm{1}(m=n_0)\sum_{l=1}^K \rho_l\rho_{k_0}  (2\pi\cdot\\
	&f\sin(2\pi f(\tau_{n,l} - \tau_{n_0,k_0})))
	\cdot \pi\alpha_l^{-\frac{1}{2}}\alpha_{k_0}^{-\frac{1}{2}}e^{-\pi^2f^2(\frac{1}{\alpha_l}+\frac{1}{\alpha_{k_0}})}.\\
	\\
	&\frac{\partial \mathcal I_{n,m,f}}{\partial \tau_{n_0,k_0}}\\
	=& \mathbbm{1}(n=n_0)\sum_{l=1}^K \rho_l\rho_{k_0}  (2\pi f\cos(2\pi f(\tau_{n_0,k_0} - \tau_{m,l})))\\
	&\cdot \pi\alpha_l^{-\frac{1}{2}}\alpha_{k_0}^{-\frac{1}{2}}e^{-\pi^2f^2(\frac{1}{\alpha_l}+\frac{1}{\alpha_{k_0}})}\\
	&+\mathbbm{1}(m=n_0)\sum_{l=1}^K \rho_l\rho_{k_0}  (-2\pi f\cos(2\pi f(\tau_{n,l} - \tau_{n_0,k_0})))\\
	&\cdot\pi\alpha_l^{-\frac{1}{2}}\alpha_{k_0}^{-\frac{1}{2}}e^{-\pi^2f^2(\frac{1}{\alpha_l}+\frac{1}{\alpha_{k_0}})}.
\end{align*}
	With the above result, we can numerically check the rank of Jacobian matrix $J(\xi) = \frac{\partial \mathcal G(\xi)}{\partial\xi}$.


\vfill
\noindent{\bf Discussion of estimating the noise variance $\sigma^2$.}\\
In the paper, we provide two ways to estimate the variance $\sigma^2$ of the noise $\varepsilon$ in the model. 
 \begin{enumerate}
     \item As it is mentioned in Section \ref{sec-test}, if $N>1$, i.e,  we can use sample variance to estimate the $\sigma^2$. That is:
 we have samples $y_{i,j}$ $\forall i=1,\dots, m, j=1,\dots,N$. Let $\bar y_i = (N)^{-1} \sum_{j=1}^Ny_{i,j}$ and $\hat\sigma^2 = (mN)^{-1}\sum_{i=1}^m\sum_{j=1}^N (y_{i,j} - \bar y_i)^2$.
    \item If $N=1$, let's assume $\varepsilon_i \sim N(0,\sigma^2)$ and $\gamma = 0$. Then we can apply Theorem III.2 to construct a consistent estimate of $\sigma^2$. Consider $\mathfrak M'\subset \mathfrak M$ and $\mathfrak r' = \text{dim}(\mathfrak M')$, $\mathfrak r = \text{dim}(\mathfrak M)$, let 
    \[\tilde T_N' = \min_{x\in\mathfrak M'}\|\hat y - x\|_2^2 ,\, \tilde T_N = \min_{x\in\mathfrak M}\|\hat y - x\|_2^2.
    \]
    Then let,
    \begin{equation}\label{eq: sigHat}
    \hat\sigma^2 = \frac{\tilde T_N' - \tilde T_N}{\mathfrak r - \mathfrak r'}.
    \end{equation}
    According to Theorem III.2, we know that under the true model $T_N'-T_N$ follows central $\chi^2$ distribution with $\mathfrak r - \mathfrak r'$ degrees-of-freedom asymptotically. Therefore $\hat\sigma^2$ is a consistent estimate of $\sigma^2$, i.e. $\hat\sigma^2\rightarrow \sigma^2$ as $\mathfrak r' - \mathfrak r\rightarrow \infty$. 
    More specifically, as mentioned in section III.C, we assume that our manifold can be decomposed to be a sum of smooth manifold and linear space. Therefore, for an $x_0\in\mathfrak M' = \mathcal M +\mathcal L'$, we can construct a linear space $\mathcal L$, s.t $L'\subset L$. Then, let $\mathfrak M = \mathcal M + \mathcal L$. we can compute eq.(\ref{eq: sigHat}).
 \end{enumerate}
 Below, we will show how to use this general strategy to construct the $\mathcal L$ in each application mentioned in the paper. The key idea is that we can always leave out some observations to construct the $\mathcal L$.
 \begin{enumerate}
 \item Matrix completion: Denote the set of observation indices as $\Omega_0$  manifold:
 $\mathfrak M' = \mathcal M_r +\mathcal L'$, where $\mathcal L' = \{X\in\mathbb R^{n_1\times n_2}: X_{i,j} = 0,\, \forall (i,j)\in\Omega_0\}$.  To estimate the $\sigma^2$, we can leave out some observation, i.e. we form a smaller observation set $\Omega_1\subset\Omega_0$. Then the new manifold is $\mathfrak M = \mathcal M_r +\mathcal L$, where $\mathcal L =\{X\in\mathbb R^{n_1\times n_2}: X_{i,j}=0, \forall (i,j) \in \Omega_1\}$. We can see that $\mathcal L'\subset \mathcal L \Rightarrow \mathfrak M'\subset \mathfrak M$. Therefore, according to eq.(\ref{eq: sigHat}), we can estimate $\sigma^2$ as following:
 \begin{align}\label{eq: matCom}
   \tilde T_N' =& \min_{X\in\mathcal M_r} \sum_{(i,j)\in\Omega_0}
    (\hat Y_{ij} - X_{ij})^2,\nonumber\\
    \tilde T_N =& \min_{X\in\mathcal M_r} \sum_{(i,j)\in\Omega_1}
    (\hat Y_{ij} - X_{ij})^2,\nonumber\\
     \hat\sigma^2 =& \frac{\tilde T_N' - \tilde T_N}{|\Omega_0| - |\Omega_1|}.
 \end{align}
\item Complex matrix completion: It is similar to real matrix completion. By leaving out some observations, we have a smaller set of observation indices $\Omega_1\subset\Omega_0 $, and 
\begin{align*}
    \mathcal L' = \{X\in\mathbb C^{n_1\times n_2}, X_{ij} = 0,\, \forall (i,j)\in\Omega_0\}\\
    \mathcal L = \{X\in\mathbb C^{n_1\times n_2}, X_{ij} = 0,\, \forall (i,j)\in\Omega_1\}
\end{align*}
Let $\tilde T_N'$ be the objective value of eq.(24) in the paper with respect to observation set $\Omega_0$ and $\tilde T_N$ be the result with respect to observation set $\Omega_1$.
Then, we can estimate the $\sigma^2$:
\[
     \hat\sigma^2 = \frac{\tilde T_N' - \tilde T_N}{|\Omega_0| - |\Omega_1|}
\]
 \item Rank-$r$ tensor completion: It is similar to matrix completion problem: Denote the manifold of rank-$r$ tensors as $\mathcal M_r$, and there is an observation index $\Omega_0$. By leaving out some observations, we have $\Omega_1\subset\Omega_0$. Let's define,
 \begin{align*}
    \mathcal L' = \{X\in\mathbb R^{n_1\times n_2}, X_{ijk} = 0,\, \forall (i,j,k)\in\Omega_0\}\\
    \mathcal L = \{X\in\mathbb R^{n_1\times n_2}, X_{ijk} = 0,\, \forall (i,j,k)\in\Omega_1\}
\end{align*}
and
\[ 
\mathfrak M' = \mathcal M_r + \mathcal L',\, \mathfrak M = \mathcal M_r + \mathcal L.
\] We can see $\mathfrak M' \subset \mathfrak M$. According to the Theorem III.2, we can construct the $\hat\sigma^2$ similar to eq.(\ref{eq: matCom}),
\begin{align*}
   \tilde T_N' =& \min_{X\in\mathcal M_r} \sum_{(i,j,k)\in\Omega_0}
    (\hat Y_{ijk} - X_{ijk})^2,\\
    \tilde T_N =& \min_{X\in\mathcal M_r} \sum_{(i,j,k)\in\Omega_1}
    (\hat Y_{ijk} - X_{ijk})^2,\\
     \hat\sigma^2 =& \frac{\tilde T_N' - \tilde T_N}{|\Omega_0| - |\Omega_1|}.
\end{align*}
 \item Demixing: It can be viewed as a tensor completion problem in our setting. The difference between the demixing problem and rank-$r$ tensor completion problem is the way of parameterizing. In the rank-r tensor completion problem, we parameterize the tensor with rank. In the demixing problem, we parameterize the tensor as the cross-correlation function of the frequency domain signals. However, in estimating $\sigma^2$, what matters is the $\mathcal L$ part, which is not related to the parameterization of the $\mathcal M$ part.      
 \item Neural networks: Suppose we have $m$ observations, i.e. $y\in \mathbb R^m$. Then we say that our set of observation indices are all the indices i.e. $\Omega_0 = \{1, 2, \dots, m\}$. Then $\mathcal L' = \{X\in\mathbb R^m: X_i = 0, \, \forall i\in\Omega_0\} = \{0\}$. By leaving out some observations, we have $\Omega_1\subset\Omega_0$, $\mathcal L = \{X\in\mathbb R^m: X_i = 0, \, \forall i\in\Omega_0\}\supset\mathcal L'$, according to the eq.(27) in the paper, $\sigma^2$ is estimated as:
 \begin{align}\label{eq:1NN}
   \tilde T_N' =& \min_{U\in\mathbb R^{d\times r}} \sum_{i=1}^m
    (y_i - \mathbf 1^\top q(U^\top x_i))^2,\nonumber\\
   \tilde T_N' =& \min_{U\in\mathbb R^{d\times r}} \sum_{i\in \Omega_1}
    (y_i - \mathbf 1^\top q(U^\top x_i))^2,\nonumber\\
     \hat\sigma^2 =& \frac{\tilde T_N' - \tilde T_N}{m - |\Omega_1|}.
 \end{align}
 \item Matrix sensing: As mentioned in the paper, matrix sensing is a special case of one-hidden-layer neural networks with quadratic activation function.
 \end{enumerate}
 Below we also present two numerical examples to show the performance of the estimate of the sigma:
 \begin{enumerate}
     \item Matrix completion:
     Table \ref{table:sim2} shows a result of estimating $\sigma^2$ for each rank $r$. In this experiment, $n_1=n_2=100$, true rank $r^*=6$, $|\Omega_0|=8000$, $\sigma=10$, $N=1$. In practise, we may not know the true rank, therefore, we compute the estimate of $\sigma^2$ for each rank $r$ ranging from $1$ to $8$. $\sigma^2$ is estimated by $\hat\sigma^2$ in eq.(\ref{eq: matCom}) with $|\Omega_1| = 2000$. When $r<r^*$, $\hat \sigma^2$ largely overestimates the $\sigma^2$ and decreases hugely as $r$ increases because part of the signal is treated as noise. When $r>r^*$, $\hat\sigma^2$ become stable since it is over-fitting the noise. We can also see that when $r = r^*$, our $\hat\sigma^2$ is close to $\sigma^2$.
     \begin{table}[h]
\caption{Estimate of $\sigma^2$ in matrix completion with true rank $r^*=6$.}
\vspace{.1in}
\centering
\begin{tabular}{|c|c||c|c|}\hline
	rank &$\hat \sigma^2$  &rank &  $\hat\sigma^2$ \\\hline
	1&34995.5  &5& 5050.63 \\\hline
	2&26751.3  &\bf 6& \bf 97.7\\\hline
	3&18719.6  &7&96.6\\\hline
	4& 11231.8 &8&96.7 \\\hline
\end{tabular}
\label{table:sim2}
\end{table}
\item Matrix sensing (One-hidden-layer neural networks with quadratic activation). Table \ref{table:MS} shows a result of estimating $\sigma^2$ for each rank $r$. In this experiment, $d=50$, true rank $r^*=3$ (the number of hidden nodes), $m=|\Omega_0|=500$, $\sigma=1$, $N=1$. We compute the estimate of $\sigma^2$ for each rank $r$ ranging from $1$ to $4$. $\sigma^2$ is estimated by $\hat\sigma^2$ in eq.(\ref{eq:1NN}) with $|\Omega_1| = 400$. We can see that our estimator $\hat\sigma^2$ is close to the true $\sigma^2$ when $r = r^*$.
\begin{table}[h]
        \caption{Estimate of $\sigma^2$ in matrix sensing ($r^* = 3$).}
        \vspace{.1in}
        \centering
        \begin{tabular}{|c|c||c|c|}\hline
            rank &$\hat \sigma^2$  &rank &  $\hat\sigma^2$ \\\hline
	        1&8952.8  &4& 1.04 \\\hline
	        2&1498.8  & 5&1.12\\\hline
	        \bf 3&\bf 1.12  &6&0.88\\\hline
        \end{tabular}
        \label{table:MS}
    \end{table}
 \end{enumerate}

\end{document}